%% file: paper.tex
\documentclass{article} %
\usepackage[final]{colm2026_conference}

\usepackage[leftcaption]{sidecap} 

\usepackage[table]{xcolor}
\usepackage{microtype}
\usepackage{hyperref}
\usepackage{url}
\usepackage{booktabs}
\usepackage{amsmath,amssymb}
\usepackage{graphicx}
\usepackage{multirow}
\usepackage{xcolor}
\usepackage{subcaption}
\usepackage{tikz}
\usetikzlibrary{positioning, fit}
\usepackage{siunitx}   

\usepackage{lineno}

\usepackage{xspace}

\newcommand{\fref}[1]{Figure~\ref{#1}}
\newcommand{\tref}[1]{Table~\ref{#1}}

\newcommand{\myparagraph}[1]{\paragraph{#1}}

\definecolor{darkblue}{rgb}{0, 0, 0.5}
\hypersetup{colorlinks=true, citecolor=darkblue, linkcolor=darkblue, urlcolor=darkblue}

\title{Cooperative Profiles Predict Multi-Agent LLM Team Performance in AI for Science Workflows}

\author{Shivani Kumar, Adarsh Bharathwaj, David Jurgens\\
\texttt{kshivan@umich.edu, adarshb@umich.edu, jurgens@umich.edu}\\
University of Michigan}

\begin{document}

\ifcolmsubmission
\linenumbers
\fi

\maketitle

\begin{abstract}
Multi-agent systems built from teams of large language models (LLMs) are increasingly deployed for collaborative scientific reasoning and problem-solving. These systems require agents to coordinate under shared constraints, such as GPUs or credit balances,  where cooperative behavior matters. Behavioral economics provides a rich toolkit of games that isolate distinct cooperation mechanisms, yet it remains unknown whether a model's behavior in these stylized settings predicts its performance in realistic collaborative tasks.
Here, we benchmark 41 open-weight LLMs across six behavioral economics games and show that game-derived cooperative profiles robustly predict downstream performance in  AI-for-Science tasks, where teams of LLM agents collaboratively analyze data, build models, and produce scientific reports under shared budget constraints. Models that effectively coordinate in games and invest in multiplicative team production (rather than greedy strategies) produce better scientific reports across three outcomes, accuracy, quality, and completeness. These associations hold after controlling for multiple factors, indicating that cooperative disposition is a distinct, measurable property of LLMs not reducible to general ability. Our behavioral games framework thus offers a fast diagnostic for screening cooperative fitness before costly multi-agent deployment.
\end{abstract}

\section{Introduction}
\label{sec:intro}

As agentic AI systems mature, multi-agent LLM teams are deployed in scientific workflows \citep{hong2024metagpt, wu2023autogen}, software development \citep{qian2024chatdev}, and organizational decision-making \citep{ Borghoff2025An}.
These agents and teams coordinate to accomplish a shared goal under common constraints, such as GPU hours or API credits, where success requires cooperative behavior. 
For humans, cooperation is critical for collective intelligence, enabling teams to achieve outcomes that exceed the capacities of any individual \citep{Momennejad2021CollectiveMSA, Shum2019TheoryOMA}. Yet, decades of research in social psychology and organizational behavior demonstrate that cooperation is not merely a coordination problem---it is shaped by group identity, role expectations, incentive structures, and trust \citep{Fiedler2023MotivatedCIA, Zolin2014HowERA, tajfel1970experiments}. Even arbitrary group labels can trigger in-group favoritism and strategic withholding of shared resources between people \citep{tajfel1971social,everett2015preferences}.
As LLMs are increasingly deployed in teams, here, we draw upon frameworks from behavioral economics to measure the cooperative profile of agentic LLMs and to test whether this profile is predictive of the eventual success in real team-based workflows.

Recent work demonstrates that LLMs placed in multi-agent settings can reproduce---and sometimes amplify---human-like coordination biases \citep{xie2024canllm, doi:10.1126/sciadv.adu9368, zhang-etal-2024-exploring}. However, existing evaluations focus primarily on dyadic interactions or single-game settings \citep{Huang2024HowFAA, Lore2024-zz, akata2023playing, xie2024canllm}, leaving gaps in both how teams of agents might cooperate and whether game-level behavioral diagnostics can predict real-world collaborative performance.

We address this gap with a two-part investigation. First, we evaluate 41 open-weight LLMs across six behavioral economics games that capture distinct cooperation challenges: coordination (Weakest-Link), commons management (Common-Pool Resource with and without sanctions), threshold collective action (Collective Risk), multiplicative team production (O-Ring), and public goods provision. Each game has established Nash equilibria and Pareto optima, providing principled references for comparing models' behaviors. Second, we deploy the same models as teams of specialized agents tasked with collaborative scientific analysis under shared budget constraints. We then assess whether models' cooperative behavior observed in the games predicts downstream task performance using Bayesian hierarchical measurement-error models. %

Our work has the following three contributions:
\textbf{First}, via the evaluation of 41 open-weight LLMs spanning multiple families and scales (0.6B–70B) across six behavioral economics games,
we reveal systematic differences in cooperative behavior, showing how models vary in coordination, resource sharing, and team production.
\textbf{Second}, behavioral game metrics credibly predict AI-for-Science outcomes. Models that exert higher effort in coordination games produce higher-accuracy scientific reports, while models that withdraw from multiplicative team production perform worse. 
\textbf{Third}, we demonstrate that behavioral games serve as an efficient proxy for evaluating agentic collaborativeness, reducing cost by orders of magnitude (median gains of 67x faster and 38x fewer tokens) enabling rapid screening of LLMs for deployment in multi-agent systems.

\section{Related work}
\label{sec:related}

Cooperation can be defined as the willingness to incur personal costs to benefit a group \citep{10.1093/qje/qjs074, doi:10.1177/1529100613496959}, and is a central challenge in both human societies and artificial multi-agent systems \citep{tomasello2009we, gunal2025exploring}. Game theory provides a formal language for studying cooperation through the lens of \emph{social dilemmas}: situations where individual and collective incentives diverge \citep{axelrod1984evolution, nowak2006five, Sadekar2025Drivers}. Five mechanisms sustain human cooperation---kin selection, direct reciprocity, indirect reciprocity, network reciprocity, and group selection \citep{nowak2006five}---yet cooperation frequently breaks down in laboratory experiments with humans, particularly in commons dilemmas and public goods settings \citep{rand2013human, ledyard1995public, Drouvelis2025Reciprocity}. As multi-agent LLM systems take on collaborative tasks that require coordination under shared constraints, the same structural tensions arise, in which agents must decide whether to invest in collective outcomes or conserve resources for individual performance \citep{dafoe2020open, dafoe2021cooperative, piedrahita2025corrupted, coopeval_iclr_2026}. %

Behavioral economics has developed a rich toolkit of experimental games to probe these tensions, each isolating a distinct cooperation mechanism. Minimum-effort (``weakest-link'') games \citep{vanHuyck1990tacit, devetag2007coordination} capture settings where team output depends on the least capable member; human groups consistently converge to inefficient low-effort equilibria despite mutual gains from coordination. Common-pool resource (CPR) experiments \citep{ostrom1990governing, cardenas2003real} study the tragedy of the commons, where individually rational extraction depletes shared resources; peer sanctioning mechanisms substantially improve sustainability \citep{casari2003sanctions, fehr2000cooperation}. Threshold public goods games, like the collective-risk social dilemma \citep{milinski2008collective, tavoni2011inequality}, model situations where groups must collectively contribute enough to avert catastrophic loss. %
 Multiplicative team production, formalized by \citeauthor{kremer1993oring}'s O-Ring theory \citep{kremer1993oring} and moral hazard in teams \citep{holmstrom1982moral}, captures settings where one low-quality contribution can destroy overall output. Finally, voluntary contribution mechanisms \citep{isaac1988group, chaudhuri2011sustaining} measure the tension between free-riding and pro-social investment across group and global pools.

Cooperative behavior in LLMs has been shown to broadly mirror human-like dynamics in strategic settings \citep{Aher2022Using, akata2023playing, Gao2023Large, li2023tom, Li2023MetaAgents, Mei2024A}, where models can act as economic agents that reproduce stylized behavioral patterns \citep{horton2023large}, exhibit generosity beyond humans in certain games \citep{fontana2024nicer}, and display deviations from Nash equilibria that resemble human decision-making \citep{brookins2024playing}. At the same time, their behavior is highly sensitive to contextual factors such as framing, incentives, and model scale, leading to substantial variability in cooperation across settings \citep{akata2023playing, Lore2024-zz, zhang2024building, Phelps_2025}. Multiple works have introduced new frameworks for benchmarking model performance on these games \citep{liang2025llm,buscemi2025fairgame}.
However, these studies focus primarily on dyadic interactions or single-game settings, and none test whether game-level behavioral diagnostics predict performance in realistic collaborative tasks.

Simultaneously, multi-agent LLM systems are increasingly applied to scientific workflows \citep{Wang2023ScientificDIA, zhang2025llm_scientific_method}. 
\citet{boiko2023autonomous} demonstrate autonomous chemical research with LLMs performing literature search, experimental planning, and robotic execution. \citet{bran2024chemcrow} augment LLMs with chemistry tools for synthesis planning and property prediction. \citet{pantiukhin2025multiagent} integrates LLMs in complex, geoscientific data repositories for accelerated data processing and collaborative problem-solving capabilities. \citet{ghafarollahi2024sciagents} automate multi-step scientific discovery through multi-agent graph reasoning. Multi-agent frameworks like MetaGPT \citep{hong2024metagpt} and AutoGen \citep{wu2023autogen} assign specialized roles to LLM agents for collaborative problem-solving, while ``The AI Scientist'' \citep{lu2026towards} attempts end-to-end automated discovery including ideation, experimentation, and paper writing.

\section{Measuring Collaborative Profiles using Behavioral Games}
\label{sec:games}

\subsection{Experimental setup}
\label{sec:games:setup}

We evaluate 41 LLMs across six behavioral economics games, each chosen to probe a distinct cooperation mechanism. The default configuration uses two groups of five players, with five rounds per simulation and 50 simulations per model. Agents receive a system prompt encoding their group membership, game rules, and payoff structure, then provide decisions in structured JSON format. In addition to the baseline condition, we test each model with an alternate prompt variant, chain-of-thought (CoT) prompting, theory-of-mind (ToM) prompting, and parameter sweeps that vary group size and game-specific payoff parameters. %
Appendix \ref{app:prompts} has the full prompt and variants. 

Two game-theoretic benchmarks anchor our analysis. A \emph{Nash equilibrium} is a strategy profile where no player can unilaterally improve their own payoff. A \emph{Pareto optimum} is an outcome where no player can be made better off without making another worse off, hence it represents the socially efficient outcome. We measure each model's behavior relative to these benchmarks, in which Pareto-proximate play indicates effective cooperation, whereas Nash-proximate play indicates strategic individualism.

\subsubsection{Game definitions}
Each game consists of a player who is assigned to a group and makes a decision to extract/contribute relative to a public pool of resources/tokens; the player utility is derived from the associated payoff function that is based on an individual value (e.g., the amount extracted) and a group-based value, which may depend on the behaviors of the group's members. We formally define each game's decision space, payoff function, and equilibrium structure. Appendix Table~\ref{tab:games} provides a summary.

\myparagraph{Weakest-Link (Minimum-Effort Coordination).} Based on \citet{vanHuyck1990tacit}. Each of the $N{=}10$ players independently selects an effort $e_i \in \{0, \ldots, 10\}$. Payoff depends on the \emph{global} minimum across \emph{all players in both groups}: $u_i = 2 \cdot \min_{j \in \mathcal{N}} e_j - e_i$. This creates a cross-group coordination challenge, in which, even if one group coordinates perfectly, a single low-effort player in the other group drags everyone's payoff down. At the same time, however, players are individually benefited from choosing lower efforts as opposed to higher ones. Any symmetric profile is a Nash equilibrium; $e^*{=}0$ is risk-dominant and $e^*{=}10$ is Pareto-dominant. Human subjects consistently converge to low-effort equilibria (mean effort ${\approx}\,3$--4) within 5--10 rounds \citep{vanHuyck1990tacit, devetag2007coordination}. For LLMs, success means sustaining high-effort coordination across both groups simultaneously, hence our primary metric is \emph{average effort}.

\myparagraph{Common-Pool Resource (CPR).} Based on \citet{ostrom1990governing}. All $N{=}10$ players share a single resource pool ($C{=}100$), in which each player extracts $x_i \in \{0, \ldots, 10\}$. Payoff is $u_i = x_i + 3 \cdot \max(0, C - \sum_{j} x_j) / N$, which means extraction yields individual gain but depletes the pool shared by both groups. However, there is also a collective gain in leaving leftover resources at the end, since that contributes to the payoff as well. Hence, each group benefits from restraining its own extraction only if the other group reciprocates, but unilateral restraint by one group can be exploited by the other. Nash equilibrium is maximum extraction ($x_i{=}10$); Pareto optimum is zero extraction. Human subjects extract 40--60\% of the socially optimal level \citep{ostrom1990governing, cardenas2003real}. For LLMs, success means restraining extraction below the selfish Nash level despite the risk of being exploited by the other group, hence our primary metric is \emph{average extraction}.

\myparagraph{CPR with Sanctioning.} Extends CPR with a sanctioning phase inspired by \citet{casari2003sanctions} and \citet{fehr2000cooperation}. After the shared extraction phase, described in CPR, players can spend tokens to sanction over-extractors \emph{within their own group}. Thus, our payoff function is $u_i^{\text{final}} = u_i^{\text{CPR}} - \sum_j s_{ij} - 2\sum_j s_{ji}$, where inflicting a specific sanction costs 1 token but being sanctioned costs 2 tokens of damage. Sanctions are only visible within groups, which means groups can police their own members but cannot penalize free-riders in the other group. Peer sanctioning reduces extraction by 30--60\% in human experiments \citep{fehr2000cooperation}. We include a sanctioning variant to test how cooperative behavior might differ in the face of explicit institutional constraints; specifically, we hypothesize that LLMs that productively respond to such constraints (more near-Pareto behavior) are more likely to collaborate in real settings in the absence of cross-group governance. For LLMs, success means both reducing extraction and strategically deploying within-group sanctions, hence our primary metric is \emph{average extraction}.

\myparagraph{Collective Risk (Threshold Public Goods).} Based on \citet{milinski2008collective}. Over $R{=}10$ rounds, each of the $N{=}10$ players contributes $c_i^{(r)} \in \{0, \ldots, 10\}$ per round from a per-round endowment. Contributions are pooled across both groups. If the \emph{global total} reaches threshold $T{=}100$, all players keep their remaining savings; otherwise, with probability $p{=}0.5$, all savings are lost. The threshold creates a collective action problem spanning both groups: each group must trust that the other will contribute its fair share. Pareto optimum is the fair share $T/(NR) \approx 1.25$ per round; Nash equilibrium (one-shot) is to free-ride. Human groups reach the threshold in 50--60\% of trials under high risk \citep{milinski2008collective, tavoni2011inequality}. For LLMs, success means contributing near the fair-share amount. Our primary metric is \emph{average contribution per round}.

\myparagraph{O-Ring Team Production.} Inspired by \citet{kremer1993oring} and \citet{holmstrom1982moral}. Each player withdraws $w_i \in \{0, \ldots, 10\}$ from a \emph{shared pool} ($R{=}200$, depleted by all $N{=}10$ players across both groups) to build individual quality $q_i = w_i/10$. Each group's production is multiplicative within the group: $P_g = (1000 \cdot (R - \sum_{j \in \mathcal{N}} w_j)/R) \cdot \prod_{i \in g} q_i$. The system succeeds only if \emph{every group's} production exceeds the threshold ($\min_g P_g \geq 50$), yielding a shared reward: $u_i = 200/N \cdot \mathbf{1}[\text{success}] - w_i$. This creates a cross-group dilemma, in which withdrawing from the shared pool to build quality depletes resources for the other group, but insufficient quality in \emph{any} group causes system-wide failure. For LLMs, success requires balancing personal investment against the shared resource cost,  analogous to multi-agent pipelines where each team's budget consumption affects downstream teams. Our primary metric is \emph{average withdrawal}.

\myparagraph{Public Goods Game.} Based on \citet{isaac1988group}. Each player allocates endowment $E{=}10$ across three channels with distinct scopes: keep ($k_i$, private), group pool ($g_i$, multiplied by 2.0 and split among the 4 members of the player's own group), and global pool ($n_i$, multiplied by 1.5 and split among all $N{=}10$ players across both groups). Payoff: $u_i = k_i + 2.0 \sum_{j \in \mathcal{G}_i} g_j / |\mathcal{G}_i| + 1.5 \sum_{j \in \mathcal{N}} n_j / N$. The three-way allocation creates a hierarchy of cooperation: the keep pool rewards only the individual (smallest multiplier but no split), the group pool rewards in-group cooperation (higher multiplier, smaller split), while the global pool rewards universal cooperation (lower multiplier, larger split). Nash equilibrium is keeping everything; Pareto optimum is contributing to the group pool ($\phi_g{=}2.0 > 1$). Human subjects contribute 40--60\% of endowment in early rounds, declining over time \citep{ledyard1995public, chaudhuri2011sustaining}. For LLMs, success means allocating to group/global pools rather than hoarding. Our primary metric is \emph{average group pool contribution}.

\subsubsection{Models}

We evaluate 41 open-weight models from twelve families spanning 0.6B to 70B parameters: Gemma~3 (1B--27B), Gemma~4 (E2B, E4B, 26B-A4B, 31B), GPT-OSS (20B), Granite~4 (3B--32B, instruct and thinking), Llama (3.1-8B, 3.3-70B), Mistral Small~3.1 (24B), Nemotron (4B--30B), OLMo~3.1 (32B instruct and thinking), Phi-4 (14B instruct and reasoning), Qwen3 (0.6B--32B with thinking variants), Qwen3.5 (4B--35B), and Qwen3.6 (27B, 35B-A3B). All models are served via vLLM. The full model list with MMLU-Pro \citep{wang2024mmlupro} and IFEval \citep{zhou2023ifeval} benchmark scores is in Appendix~\ref{app:models}. 

\subsubsection{Metrics and Evaluation}

For each model, we compute the primary behavioral metric for each game, averaged across all simulations and rounds: average effort (Weakest-Link), average extraction (CPR and CPR+Sanction), average contribution (Collective Risk), average withdrawal (O-Ring), and average group pool contribution (Public Goods). Together, these six metrics form the model's \emph{cooperative profile}, which acts as a behavioral fingerprint that captures how the model navigates distinct cooperation challenges.

To quantify what drives cooperative behavior, we fit a Bayesian hierarchical regression predicting \emph{Pareto proximity}---the normalized distance from the Pareto optimum ($0 =$ Pareto, $1 =$ Nash)---from model characteristics, prompting strategies, and game structure. A Bayesian model lets us propagate uncertainty from the 50 sampled simulations into the regression coefficients, rather than collapsing each game to a point estimate as OLS would require. The model includes log$_{10}$(model size), a thinking-variant indicator, binary chain-of-thought (CoT) and theory-of-mind (ToM) prompting indicators, group size, game fixed effects, and model nested within family as a random intercept, estimated on 89,475 simulation-level observations across 41 models, 6 games, and multiple experimental conditions ($R^2 = 0.484$; with zero divergences). To ensure that any effects are not due to the general prompt structure, we conducted a robustness test using an alternative prompt design, described in Appendix~\ref{app:prompt_invariance}, showing no effect from the template.

\subsection{Results and discussion}
\label{sec:games:results}

\input{tables/tab_pareto_regression_compact.tex}

Model size is the dominant predictor of collaborativeness: each unit increase in $\log_{10}(\text{size})$ reduces Pareto proximity by $0.051$. Thinking variants are not further from Pareto on average, which differs from recent work suggesting reasoning models are more likely to strategize and free-ride to increase their individual gains \citep{piedrahita2025corrupted}. 
Games themselves vary enormously: Weakest-Link is the easiest game (64\% near-Pareto) while models are far less likely to behave globally optimal for CPR (0.5\% near-Pareto) and Public Goods (2.4\%). Model families also vary meaningfully and explain variance beyond size: Gemma~4 and OLMo are the most cooperative families, while Granite~4 and Nemotron are furthest from Pareto; Appendix \fref{fig:nash_pareto} visualizes models' behaviors.

Among prompting strategies, \textbf{Theory of Mind has a credible effect}. ToM-prompted agents play closer to the Pareto optimum, making it the strongest strategy effect after model size. In contrast, chain-of-thought prompting has no effect. Group size has no credible effect after controlling for game fixed effects; though, we note that only a relatively narrow range of group sizes were evaluated. Table~\ref{tab:pareto_regression} reports the full model coefficients.

\section{AI for Science experiments}
\label{sec:ailab}

To test whether cooperative behavior in games generalizes to real-world collaboration, we construct a multi-agent AI-for-Science benchmark involving data analysis and report generation. The task simulates realistic workflows with role specialization, shared budgets, and interdependent stages. We use this setup to evaluate whether game-based behavioral profiles predict downstream team performance.

\subsection{Experimental setup}
\label{sec:ailab:setup}

We design a collaborative AI-for-Science task that requires teams of LLM agents to jointly analyze a dataset and produce a scientific report. This task captures the key features of real multi-agent scientific workflows: role specialization, sequential dependencies, shared resource constraints, and quality evaluation against ground truth.

\myparagraph{Science tasks.}
We implement four realistic domain-specific tasks, each with pre-generated synthetic datasets and deterministic ground truth: (1) \textbf{
Ecology}: Multi-species population dynamics with inter-species interactions and regime shifts, (2) \textbf{Anomaly}: Multi-sensor anomaly detection with point, contextual, and collective anomalies, (3) \textbf{Biomarker}: Clinical biomarker discovery with patient subtypes and differential treatment response, and (4) \textbf{Geospatial}: Spatial resource mapping requiring spatial regression and cross-validation. Since each task has a known ground truth, we are able to precisely measure the report's accuracy.

\myparagraph{Team structure.}
Each simulation deploys four sequential teams, each composed of 2--3 specialized agents with unique roles (see Appendix~\ref{sec:role-descriptions} for descriptions). Teams work iteratively to read the artifacts produced by upstream teams and produce new artifacts for use by downstream teams. Teams may request additional work by earlier teams or ask them to re-do work deemed unacceptable.  Agents within each team have access to a Python sandbox (with pandas, scikit-learn, scipy, matplotlib), artifact reading/writing tools, and inter-team communication tools. The writing team has no code tools and must synthesize from upstream artifacts. The team implementation uses LangGraph \citep{langgraph2024} with the model's native tool calling format to better simulate real-world agentic workflows. 

\myparagraph{Budget mechanism.}
All teams share a single credit pool with total budget $B$ (varied across conditions). Token usage is charged at fixed rates per 1K input/output tokens, simulating real-world API usage (\$0.0005 per 1K input tokens, \$0.003 per 1K output). We vary budget visibility with three settings: (1) \textbf{Full}: All agents see the current budget status (spent, remaining); (2) \textbf{Team-only}: Agents see only their own team's spending; and (3) \textbf{Hidden}: Agents receive no budget information.
When budget reaches 30\% remaining, a warning is added to the prompt to finish concisely. At 15\%, agents are instructed to immediately finalize their work. Budget exhaustion results  in incomplete artifacts.

\subsubsection{Experimental conditions}

We sweep over the following variables: \textbf{Budget}: \$0.50, \$1.00, \$2.50, \$5.00, \$10.00; \textbf{Visibility}: full, team-only, hidden; \textbf{Team size}: 2 or 3 agents per team; \textbf{Tasks}: all four science tasks; \textbf{Prompting}: standard and ToM variants.
This setup yields up to 240 unique conditions per model where we run 5 simulations per condition. Running all 49,200 simulations is computationally prohibitive, so we sample 3,473 conditions to run (17,388 total simulations). We randomly sample across each variable except for team size and prompting, which were biased toward 2 agents and the standard (non-ToM) prompt due to computational efficiency.

\subsubsection{Evaluation}

We evaluate each simulation's output along three dimensions: 
\textbf{Completeness} ($\in [0,1]$): Fraction of the seven required report sections present with substantive content (abstract, introduction, dataset description, methods, results, evaluation, conclusion).
\textbf{Quality} ($\in [0,1]$): Task-agnostic heuristic score based on the presence of: (1) specific numerical results, (2) standard methodology references, (3) statistical tests, and (4) cross-references to other teams' work.
\textbf{Accuracy} ($\in [0,1]$): Task-specific score evaluating whether key domain findings are reported. For example, the ecology task checks whether the correct number of species is mentioned, whether the regime shift is detected, and whether inter-species interactions are discussed. We use \texttt{gpt-5.4-mini} as a judge for all three metrics, validating ratings with \texttt{gemini-3.1-flash-lite-preview} as a second judge; Appendix \ref{app:judge_prompts} provides full details.

We link behavioral game profiles to AI-for-Science outcomes. Each observation is a model--condition pair (model $\times$ task $\times$ budget $\times$ visibility $\times$ team size $\times$ ToM). %
We use 3,473 config/model-level variations across the condition sweep.

\myparagraph{Identification challenge.} Because game metrics are constant within model, our key predictors vary only between models. A latent capability confounder could drive both game performance and AI-for-Science outcomes: more capable models may simultaneously score higher on behavioral games and produce better scientific reports. We therefore frame our analysis as \emph{predictive association}, not causal identification, and include instruction-following capability (IFEval) as a control to test whether game metrics predict beyond general ability.

\myparagraph{Bayesian hierarchical measurement-error model.}
Standard regression treats per-model game metric averages as fixed and known. In practice, each model's game metric is estimated from 50--350 per-simulation observations with non-trivial sampling variability (standard errors of 0.0001 to 0.253 in $z$-scored units). Treating these noisy estimates as exact predictors risks both attenuation bias (coefficients biased toward zero) and overfitting to measurement noise. We address this with a two-level Bayesian hierarchical model:

\emph{Level 1 (measurement model):} For each model $i$ and game $g$, the observed metric mean $\bar{y}_{ig}$ is modeled as a noisy draw from a latent true behavioral disposition $\theta_{ig}$. Thus,
$\bar{y}_{ig} \sim \mathcal{N}(\theta_{ig},\; \mathrm{SE}_{ig}^2)$,
where $\mathrm{SE}_{ig} = s_{ig} / \sqrt{n_{ig}}$ is the estimated standard error computed from the per-simulation standard deviation $s_{ig}$ and the number of simulations $n_{ig}$. The latent dispositions are given a hierarchical prior with non-centered parameterization for sampling efficiency. $\theta_{ig} = \mu_g + \sigma_g \cdot z_{ig}, \quad z_{ig} \sim \mathcal{N}(0, 1)$,
where $\mu_g$ and $\sigma_g$ are game-level hyperparameters governing the population distribution of behavioral dispositions across models. Models with high measurement noise (large $\mathrm{SE}_{ig}$) have their $\theta_{ig}$ estimates shrunk toward the game mean $\mu_g$, while precisely measured models retain their observed values.

\emph{Level 2 (prediction model):} For model $i$ in AI-for-Science condition $k$, $
    y_{ik} \sim \mathcal{N}\!\left(\alpha + \boldsymbol{\beta}^\top \boldsymbol{\theta}_i + \boldsymbol{\gamma}^\top \mathbf{x}_k + \boldsymbol{\delta}^\top \mathbf{b}_i,\; \tau^2\right)$,
where $\boldsymbol{\beta}$ captures the effect of latent game dispositions on AI-for-Science outcomes, $\boldsymbol{\gamma}$ captures condition-level controls (log budget, task dummies, visibility dummies, team size, and ToM indicator), $\boldsymbol{\delta}$ captures model-level benchmark controls (IFEval, MMLU-Pro, MATH-500, log model size, and thinking variant indicator, all z-scored), and $\tau$ is the residual standard deviation. All game features are $z$-scored before analysis.

Priors are weakly informative: $\boldsymbol{\beta} \sim \mathcal{N}(0, 1)$, $\alpha \sim \mathcal{N}(0.5, 0.5)$, $\boldsymbol{\gamma}, \boldsymbol{\delta} \sim \mathcal{N}(0, 0.5)$, $\tau \sim \text{Half-Normal}(0.5)$, $\mu_g \sim \mathcal{N}(0, 2)$, $\sigma_g \sim \text{Half-Normal}(2)$. We fit the model using NUTS \citep{hoffman2014no} via NumPyro \citep{phan2019composable} with 4 chains, 4,000 warmup steps, and 8,000 samples per chain (32,000 total posterior draws). We report posterior means, 95\% credible intervals, and posterior probabilities of direction $P(\beta > 0)$.

This approach offers three advantages over OLS. First, it \emph{propagates measurement uncertainty}: game metrics estimated from few or highly variable simulations receive wider credible intervals rather than entering the regression as point estimates. Second, the hierarchical prior provides \emph{partial pooling}, shrinking extreme model-level estimates toward the population mean. Third, the joint estimation allows the AI-for-Science likelihood to \emph{inform} the latent game dispositions, producing more coherent inferences than a two-stage plug-in approach.

\myparagraph{Specifications.} We report a \emph{benchmark-controlled model} that includes five model-level capability controls (z-scored performance on three benchmarks, IFEval, MATH-500, and MMLU-Pro; log$_{10}$(size); and a thinking-variant indicator), condition-level controls (log(budget), task dummies, visibility dummies, team size as a categorical factor with team size 2 as reference, and a ToM indicator), and the means of all six game metrics as predictors. The benchmark scores allow us to control for relative model capabilities (e.g., math ability) that might influence outcomes.

\subsection{Results}
\label{sec:ailab:results}

\input{tables/tab_ailab_regression_compact.tex}

Predicting the success metrics on AI-for-science workflows from a model's cooperative profile shows that resource conservation (O-Ring) is the most consistently predictive game behavior, while the CPR vs CPR+Sanction contrast reveals institutional sensitivity as a key differentiator, as shown in \tref{tab:regression}. All regression parameters converged well ($\hat{R} \leq 1.01$, zero divergent transitions across all three outcome models). We summarize the main themes below.%

\textbf{O-Ring withdrawal} is negatively associated with all three metrics. This game rewards balancing consuming shared resources to build individual quality while leaving sufficient resources for all teams to succeed, and models who withdraw more produce worse scientific output across all dimensions. The effect is strongest for completeness, where a single disengaged team can prevent the pipeline from finishing.

\textbf{Weakest-Link effort} is positively associated with Accuracy but negatively with Completeness and Quality. This game's payoff for team output depends on the minimum individual contribution and models that cooperate more produce more accurate scientific reports, but in the process they complete fewer  stages and generate lower quality write-ups.

\textbf{CPR extraction} shows a credible negative effect on accuracy but is positively associated with quality. Models that extract aggressively from shared resources produce reports that are more likely to be rated higher quality---consistent with a throughput advantage where resource-hungry models generate more output at each pipeline stage.
However, \textbf{CPR+Sanction extraction} shows that  models' responsiveness to institutional constraints---i.e., the presence of sanctioning abilities---produces more-accurate and more-complete output but lower-quality reports. Models continue extracting aggressively \emph{despite} the availability of peer sanctioning, suggesting that, unlike unsanctioned CPR, models' sensitivity to punishment, not extraction level itself, is the relevant behavioral signal.

\textbf{Capability controls} capture meaningful variation with respect to model abilities. IFEval is a positive predictor for all three outcomes, confirming that instruction-following ability contributes to multi-agent task success independently of cooperative dispositions. MMLU-Pro shows a negative effect on quality  and completeness but a negative effect on  accuracy: after controlling for IFEval and game metrics, higher knowledge-benchmark performance is associated with worse task outcomes on these dimensions. This negative coefficient likely reflects a suppression effect: MMLU-Pro generally correlates with model size, which is already partially captured by log$_{10}$(size) and game metrics (cf.~\tref{tab:models}). MATH-500 shows no credible effect on accuracy but is positively associated with quality and, most strongly, completeness, suggesting that mathematical reasoning ability primarily supports the successful execution and completeness of complex scientific workflows. Thinking variants reduce accuracy and completeness but have no credible effect on quality. Model size is positively associated with quality and completeness, but has no credible effect on accuracy.

\textbf{Condition controls.} Team size (2 vs.\ 3) has a credible improvement in Accuracy and Quality, suggesting that adding a third agent teammate slightly improves report content but reduces the chance the pipeline finishes. Surprisingly, theory-of-mind prompting has a credible positive effect only on completeness, in contrast to its clear effect on cooperative game behavior (\tref{tab:pareto_regression}). This ToM result diverges from other recent work that has suggested ToM is beneficial for cooperation in multi-agent systems \citep{li2023tom,riedl2026emergent}; however, prior work has used simpler tasks (guessing games) for demonstrating ToM benefits, whereas our full AI-for-Science task is substantially more complex, demanding models reason about an entirely different task while cooperating with various groups and completing different phases of the workflow. 
Indeed, parallel work on ToM in embodied settings has shown that while models may have high literal ToM estimated through Q\&A settings, many models fail to have functional ToM in more complex, embodied settings \citep{juneja2026enacttom}; we hypothesize that a similar trend is being seen in practice here where models have more limited ToM abilities in our task-oriented setting, which hinders the potential downstream benefits of ToM.
Our results point to a need for future work to examine ToM in more complex task spaces. Budget, visibility, and task effects are reported in the full table (Appendix~\ref{app:exp2_full}).

\textbf{Computational Efficiency.} A practical motivation for predicting AI-for-Science outcomes from behavioral games is \emph{cost}: running the full multi-agent pipeline is expensive, and, ideally, behavioral games offer a fast, reusable diagnostic for assessing whether an LLM is suitable for a multi-agent workflow. While our games involve more agents and potentially more deliberation, the majority of their tokens are in the prompt (representing the previous rounds) and each agent's output is decoded in parallel from this same prompt; with modern LLM-servers like vLLM and prefix caching, each behavioral game's rollout is computationally efficient, compared with the highly sequential AI-for-Science experiments. The median time for one sweep of all six games is 1.5 minutes vs. 10.4 minutes for a single AI-for-Science experiment run. Reliable estimates of the cooperative behavior metrics can be made with $\approx$20 simulations (Appendix \fref{fig:convergence}), suggesting accurate, reusable estimates for a model can be obtained in the same time as $\sim$3 AI-for-Science experiments.

\section{Conclusion}
\label{sec:conclusion}

Effective agentic workflow requires teams of agents to productively collaborate, especially in the face of cost or resource constraints. Here, we introduce a new framework for diagnosing and predicting cooperative behavior in LLM agent teams using behavioral economic games to quantify strategic behavior. Using 41 open weight models, we demonstrate that LLMs have different cooperative behavior profiles that are only partially explained by size and capability.  Using realistic workflows for four AI-for-Science tasks, we demonstrate that an LLM's cooperative profile is predictive of tasks' outcomes (completeness, quality, and accuracy). Our results show that models that coordinate on high effort produce more accurate scientific reports, while those that withdraw from joint processes perform worse across all outcomes. Unlike prior work, we find that while theory-of-mind prompting improves cooperative behavior in games, it only improves AI-for-Science task completeness (but not quality or accuracy), highlighting a divergence between stylized cooperation and realistic task performance. Our framework offers an effective method for characterizing models and raises new possibilities for how to train or post-train models to maximize their collaborative behavior. Code and data are available at \url{https://github.com/davidjurgens/agent-cooperation}.

\section*{Ethics statement}

This work studies the cooperative behavior of LLM agents in controlled experimental settings. All experiments use open-weight models on local compute infrastructure; no human subjects are involved. The behavioral games and AI-for-Science tasks use synthetic data with no personally identifiable information. We note that understanding LLM cooperation patterns has implications for deployment: models that exhibit strong in-group bias or free-riding tendencies in games may behave similarly in real-world multi-agent systems, potentially affecting fairness and efficiency.

\bibliography{references_paper,colm2026_conference}
\bibliographystyle{colm2026_conference}

\appendix

\section{Disclosure of LLM Use}

LLMs were used throughout the research process to (i) assist in literature review, (ii) assist in developing the software for executing the experiments, particularly for debugging many vLLM issues when LLMs have different tool parsers, (iii) analyze intermediate results, and (iv) revise the text and figure layouts. LLMs include ChatGPT, Gemini, AI2 Asta, and Claude Code, with different/multiple models being used for different steps. We also use LLM-as-judge for scoring the generated reports, using \texttt{gpt-5.4-mini} as the primary judge, which is validated against another frontier-model judge, \texttt{gemini-3.1-flash-lite-preview}. Authors were heavily involved in all steps and reviewed all LLM outputs prior to execution or inclusion in the paper to ensure faithfulness and accuracy.

\section{Behavioral Game Experiment Details}
\label{app:prompts}

This appendix provides the complete prompt templates used in our behavioral game experiments. All parameter values shown correspond to the default configuration (2 groups $\times$ 5 players, endowment = 10).

\subsection{System prompt structure}

Each agent receives a system prompt assembled from the following components. The \textbf{standard} variant uses a formal numbered style:

\begin{quote}
\small\ttfamily
You are Player \{player\_id\} in \{group\_id\}.

GAME RULES:\\
\{game\_description\}

\{strategy\_instructions, if any\}

IMPORTANT: Respond ONLY with valid JSON matching the required schema. Do not include any explanation or text outside the JSON.
\end{quote}

The \textbf{alternate} variant uses a prose style for robustness testing:

\begin{quote}
\small\ttfamily
You are participating in a group decision-making study. You have been assigned as \{player\_id\} and belong to \{group\_id\}.

About this activity:\\
\{game\_description\}

\{strategy\_instructions, if any\}

Please provide your response as a JSON object matching the required format. Do not include any additional text or explanation.
\end{quote}

\subsection{Decision prompt template}

Each round, agents receive a decision prompt. The \textbf{standard} variant:

\begin{quote}
\small\ttfamily
=== Round \{round\_num\} of \{total\_rounds\} ===

PREVIOUS ROUNDS:\\
--- Round \{N\} ---\\
\{round\_summary\}

GROUP DELIBERATION (your group's discussion before this decision):\\
{[}\{player\_id\}{]}: \{message\}

Now make your decision. Your total allocation must equal exactly \{endowment\} tokens.\\
Respond with EXACTLY this JSON format (fill in integer values):\\
\{schema\_example\}
\end{quote}

\subsection{Deliberation prompt template}

When deliberation is enabled, agents discuss strategy before deciding:

\begin{quote}
\small\ttfamily
This is the GROUP DELIBERATION phase. You can discuss strategy with your group members before making your actual decision.

PREVIOUS ROUNDS:\\
\{history\}

GROUP CHAT SO FAR:\\
{[}\{player\_id\}{]}: \{message\}

Share your thoughts with your group. What strategy do you think the group should adopt? Respond with a short message (1--3 sentences).
\end{quote}

\subsection{Game descriptions}

Below are the exact game descriptions shown to agents for each of the six games used in our analysis, summarized in \tref{tab:games}.

\begin{table}[t]
\centering
\small
\begin{tabular}{lp{2.2cm}ccp{3.5cm}}
\toprule
\textbf{Game} & \textbf{Decision} & \textbf{Nash} & \textbf{Pareto} & \textbf{Cooperation challenge} \\
\midrule
Weakest-Link & Effort $[0,10]$ & 0 & 10 & Coordination on high effort \\
CPR & Extract $[0,10]$ & 10 & 0 & Restraint in commons \\
CPR+Sanction & Extract + sanction & 10,0 & 0,-- & Costly norm enforcement \\
Collective Risk & Contribute per round & 0 & $\approx$1.25/rnd & Threshold collective action \\
O-Ring & Withdraw $[0,10]$ & 0 & 6 & Multiplicative production \\
Public Goods & 3-way allocation & Keep all & Group pool & Free-riding vs. provision \\
\bottomrule
\end{tabular}
\caption{Summary of six behavioral economics games. Each probes a distinct cooperation mechanism. ``Nash'' and ``Pareto'' columns indicate the risk-dominant Nash equilibrium and the socially optimal outcome, respectively.}
\label{tab:games}
\end{table}

\subsubsection{Weakest-Link (Minimum-Effort Coordination)}

\textbf{Standard:}
\begin{quote}
\small\ttfamily
Weakest-Link (Minimum-Effort) Coordination Game.

There are 2 groups of 5 players each (10 players total).\\
Each player independently chooses an effort level from 0 to 10.

Your payoff depends on two things:\\
1. The MINIMUM effort chosen across ALL 10 players (the `weakest link')\\
2. Your own effort level

Payoff = 2 x (minimum effort across all players) - 1 x (your effort)

Higher effort is rewarded only if EVERYONE else also chooses high effort. If anyone chooses low effort, the minimum drops and high-effort players lose out.\\
The best outcome for everyone is if all players choose 10, but this requires trusting that no one will choose lower.
\end{quote}

\textbf{Alternate:}
\begin{quote}
\small\ttfamily
You are participating in a coordination activity with 7 other participants, organized into 2 groups of 5.

Each participant selects an effort level between 0 and 10. The key feature of this activity is that everyone's outcome depends on the lowest effort chosen by any single participant across all groups. This represents a `weakest link' --- the group is only as strong as its least contributing member.

Your earnings are calculated as 2 times the lowest effort chosen by anyone, minus 1 times your own effort. Choosing high effort is beneficial when others do the same, but costly if someone else chooses low effort. The challenge is coordinating on a high effort level when you cannot be certain what others will choose.
\end{quote}

\subsubsection{Common-Pool Resource (CPR)}

\textbf{Standard:}
\begin{quote}
\small\ttfamily
Common-Pool Resource Game.

There is a shared resource pool with 100 tokens.\\
Each player can extract between 0 and 10 tokens from the pool.\\
After all extractions, the remaining pool is multiplied by 3.0 (sustainability factor) and split equally among all 10 players.

Your payoff = tokens you extracted + your share of the regenerated pool.\\
Over-extraction depletes the pool for everyone.
\end{quote}

\textbf{Alternate:}
\begin{quote}
\small\ttfamily
You are participating in a shared resource activity. There is a common pool containing 100 tokens that all 10 participants share.

Each participant may withdraw between 0 and 10 tokens from the pool. After everyone has made their withdrawal, the tokens remaining in the pool are multiplied by 3.0 (representing natural replenishment) and the result is divided equally among all participants.

Your earnings are the tokens you withdrew plus your share of the replenished pool. Keep in mind that the more everyone withdraws, the less remains for replenishment, which affects everyone's share.
\end{quote}

\subsubsection{CPR with Internal Sanctioning}

\textbf{Standard:}
\begin{quote}
\small\ttfamily
Common-Pool Resource Game with Sanctioning.

PHASE 1 - Extraction:\\
There is a shared resource pool with 100 tokens.\\
Each player can extract between 0 and 10 tokens.\\
Remaining pool is multiplied by 3.0 and split among all 8 players.

PHASE 2 - Sanctioning:\\
After seeing extractions, you can spend tokens to sanction other players.\\
Each sanction token costs you 1.0 but reduces the target's payoff by 2.0.

Final payoff = extraction + sustainability share - sanction costs - sanction impacts received.
\end{quote}

\textbf{Phase 2 sanctioning prompt:}
\begin{quote}
\small\ttfamily
SANCTIONING PHASE:\\
Here are the extractions from Phase 1:\\
- \{player\_id\}: extracted \{X\}, phase 1 payoff: \{Y\}

Resource pool: 100 capacity, \{total\} extracted, \{remaining\} remaining.

Your current phase 1 payoff is \{my\_payoff\}.\\
You may now spend tokens to sanction other players in your group who over-extracted.\\
Each sanction token costs you 1.0 but reduces the target's payoff by 2.0.

Respond with JSON: \{"sanctions": \{"player\_id": <integer>, ...\}\}
\end{quote}

\subsubsection{Collective Risk (Threshold Public Goods)}

\textbf{Standard:}
\begin{quote}
\small\ttfamily
Collective-Risk Social Dilemma (Climate Game).

Over 10 rounds, all 10 players must collectively contribute at least 100 tokens to avoid catastrophe.\\
Each round, you have 10 tokens and decide how many to contribute.\\
Tokens you don't contribute are saved.

At the end of round 10:\\
- If the group total meets the threshold (100), everyone keeps their savings.\\
- If not, there is a 50\% chance everyone loses ALL their savings.

The challenge: contribute enough collectively without sacrificing too much individually.
\end{quote}

\textbf{Round status update:}
\begin{quote}
\small\ttfamily
COLLECTIVE RISK STATUS (Round \{N\} of 10):\\
Cumulative contributions so far: \{total\} / 100 (\{pct\}\%)\\
Still need \{remaining\} more tokens across all players to avoid catastrophe.

THIS IS THE FINAL ROUND. If the threshold is not met, there is a 50\% chance ALL savings are lost.

You have 10 tokens this round. Decide how many to contribute.\\
Respond with JSON: \{"contribute": <integer>\}
\end{quote}

\subsubsection{O-Ring Team Production}

\textbf{Standard:}
\begin{quote}
\small\ttfamily
O-Ring Team Production Game.

There are 2 teams of 5 workers each (10 workers total), sharing a common resource pool of 200 units.

Each worker withdraws between 0 and 10 units from the pool for productive effort. Withdrawals are costly: you pay 1.0 per unit withdrawn.

Team production is MULTIPLICATIVE (O-Ring): each member's quality (withdrawal / 10) multiplies together. If anyone on your team contributes nothing, your team produces nothing --- like the Space Shuttle Challenger, where one faulty O-ring destroyed the mission.

Additionally, all withdrawals deplete the shared pool. The fraction of pool remaining scales all teams' production equally.

Team\_t production = 1000.0 x (pool remaining / 200) x product of all team member qualities

The system succeeds ONLY if EVERY team's production >= 50. If the system succeeds, a reward of 200.0 is split equally among all 10 workers.

Your payoff = (200.0 / 10 if system succeeds, else 0) - 1.0 x your withdrawal
\end{quote}

\subsubsection{Public Goods Game}

\textbf{Standard:}
\begin{quote}
\small\ttfamily
Intergroup Public Goods Game.

You have 10 tokens to allocate across three options:\\
1. KEEP: Token stays with you.\\
2. GROUP POOL: Multiplied by 2.0 and split equally among your group (8 members).\\
3. GLOBAL POOL: Multiplied by 1.5 and split equally among ALL players (10 total).

Your total allocation must equal exactly 10 tokens.
\end{quote}

\textbf{Alternate:}
\begin{quote}
\small\ttfamily
You are taking part in a group investment activity. You have been given 10 tokens that you must distribute across three options.

You may keep tokens for yourself, receiving their full value. You may place tokens into your group's shared pool. The group pool is multiplied by 2.0 and the result is divided equally among all 5 members of your group. You may also place tokens into a global pool that benefits everyone. The global pool is multiplied by 1.5 and divided equally among all 10 participants.

You must distribute all 10 tokens with none left over.
\end{quote}

\subsection{Group Size Parameter Sweeps}

Behavioral games all used two groups with a parameter sweep value for the size of the groups as shown in \tref{tab:group_sizes}. Both groups had the same size. Fewer group sizes were used for CPR+Sanction as it was intended as a variant of CPR designed to test the impact of sanctioning.

\begin{table}[th]                                                         \centering                                                           
  \begin{tabular}{ll}                                                  \toprule \textbf{Game} & \textbf{Group sizes} \\                     
  \midrule                                                             
  Weakest Link    & 3, 5, 8, 10 \\ 
  Collective Risk & 3, 5, 8, 10 \\ 
  O-Ring Team     & 3, 5, 8 \\  
  Public Goods    & 3, 4, 5, 8, 10 \\  
  CPR             & 3, 5, 8, 10 \\
  CPR+Sanction    & 5 \\                                               
  \bottomrule                                                          
  \end{tabular}                                                        
  \caption{Group sizes (agents per group) used in each game. All games use 2 groups.}                    
  \label{tab:group_sizes}                                                                                                                                                                      
  \end{table} 

\section{AI-for-Science agent prompts}
\label{app:ailab_prompts}

Below is the system prompt template used for all AI-for-Science agents, followed by representative role descriptions.

\subsection{System prompt template}

\begin{quote}
\small\ttfamily
You are \{role\_name\}, a member of the \{team\_display\_name\} team.

PROJECT: \{science\_task\_description\}

YOUR ROLE: \{role\_description\}

TEAM MEMBERS:\\
- \{member.name\} (\{member.id\})\\
\{...\}

\{budget\_status\}

BUDGET RULES:\\
- Every tool call and LLM response costs credits from the shared budget.\\
- The project FAILS if budget runs out before the final report is complete.\\
- If remaining budget is below 30\%: finish your current task concisely and call mark\_complete. Skip exploratory or speculative work.\\
- If remaining budget is below 15\%: call mark\_complete immediately with whatever you have. Do NOT start new analyses.\\
- Keep code short and targeted. Avoid redundant runs or printing excessive output.

TOOLS:\\
- run\_code(code, description): Execute Python code. The dataset is pre-loaded as `df'. Available libraries: pandas, numpy, scikit-learn, scipy, matplotlib, seaborn. Timeout: 30s. Memory limit: 2GB.\\
- read\_artifact(name): Read a named artifact from the workspace.\\
- write\_artifact(name, content, description): Save an artifact for other teams.\\
- list\_artifacts(): List all available artifacts.\\
- send\_message(to\_team, content): Send a message to another team.\\
- request\_additional\_work(target\_team, request): Request another team to produce specific NEW output.\\
- request\_team\_redo(target\_team, request, justification): Request another team to REDO their work from scratch. EXPENSIVE.\\
- mark\_complete(summary): Signal your work is done.

ARTIFACTS:\\
- data team $\rightarrow$ ``dataset\_description''\\
- model team $\rightarrow$ ``model\_architecture''\\
- eval team $\rightarrow$ ``evaluation\_report''\\
- writing team $\rightarrow$ ``final\_paper''

Before starting work, use read\_artifact to review artifacts from upstream teams.

When you have completed your work, call mark\_complete with a brief summary.
\end{quote}

\subsection{Representative role descriptions}
\label{sec:role-descriptions}

\textbf{Data Lead:}
\begin{quote}
\small\ttfamily
You plan the data analysis strategy. Write data loading and cleaning code, handle missing values, and produce a thorough dataset\_description artifact that summarizes the data structure, quality issues, and key patterns found.
\end{quote}

\textbf{ML Engineer:}
\begin{quote}
\small\ttfamily
You plan the modeling approach. Design the ML pipeline, select appropriate algorithms, write the primary model training code, and produce a model\_architecture artifact describing your approach and rationale.
\end{quote}

\textbf{Eval Lead:}
\begin{quote}
\small\ttfamily
You plan the evaluation strategy. Design comprehensive metrics, write evaluation code, and produce an evaluation\_report artifact summarizing model performance with statistical rigor.
\end{quote}

\textbf{Lead Author} (no code tools):
\begin{quote}
\small\ttfamily
You plan the paper structure and produce the final\_paper artifact. Synthesize all team outputs into a coherent scientific report with proper sections: abstract, introduction, dataset description, methods, results, evaluation, and conclusion. Edit for clarity and flow.
\end{quote}

\subsection{Science task descriptions}

Each task description is injected into the system prompt via the \texttt{\{science\_task\_description\}} variable:

\textbf{Ecology:}
\begin{quote}
\small\ttfamily
Analyze a multi-species ecological population dynamics dataset. Identify inter-species relationships, detect regime shifts, and build predictive models for future population sizes.
\end{quote}

\textbf{Anomaly Detection:}
\begin{quote}
\small\ttfamily
Build an anomaly detection system for multi-sensor industrial data. Detect point, contextual, and collective anomalies across 8--12 sensors.
\end{quote}

\textbf{Biomarker Discovery:}
\begin{quote}
\small\ttfamily
Analyze a clinical biomarker dataset to predict treatment response, discover patient subtypes, and identify biomarkers for differential effects.
\end{quote}

\textbf{Geospatial Mapping:}
\begin{quote}
\small\ttfamily
Build spatial prediction models for resource concentration mapping. Compare spatial vs.\ non-spatial approaches using spatial cross-validation.
\end{quote}

\subsection{Planning and work prompts}

\textbf{Planning prompt} (issued to team lead at the start of each phase):
\begin{quote}
\small\ttfamily
You are \{role\_name\}, the team lead for \{team\_display\_name\}.

Review any available artifacts from upstream teams and consider the remaining budget. Then assign specific tasks to each team member.

Produce a brief work plan (3--5 bullet points) describing what each team member should do in this phase.
\end{quote}

\textbf{Work prompt} (issued to each agent with their assigned task):
\begin{quote}
\small\ttfamily
YOUR TASK: \{specific\_task\}

The team plan is:\\
\{team\_plan\}

\{If code tools available:\} Use run\_code to execute Python code for your analysis. Write results to artifacts using write\_artifact.\\
\{If no code tools:\} Produce your contribution as text. Use write\_artifact to save your output.

When done, call mark\_complete with a brief summary.
\end{quote}

\subsection{Additional agent roles}

When team size is 3, each team gains a third specialist:

\textbf{Domain Expert} (Data team):
\begin{quote}
\small\ttfamily
You validate feature relevance and write feature engineering code. Assess which features are scientifically meaningful, create derived features, and document feature engineering decisions.
\end{quote}

\textbf{Experimentalist} (Model team):
\begin{quote}
\small\ttfamily
You run ablation studies and model comparisons. Write code to compare multiple approaches (baselines, feature subsets, hyperparameters), produce an experiment\_results artifact with quantitative comparisons.
\end{quote}

\textbf{Error Analyst} (Evaluation team):
\begin{quote}
\small\ttfamily
You analyze model errors and failure modes. Write code for confusion matrices, error distributions, subgroup performance analysis, and identify systematic failure patterns.
\end{quote}

\textbf{Results Writer} (Writing team, no code tools):
\begin{quote}
\small\ttfamily
You write the results, discussion, and conclusion sections. Reference actual numbers from the evaluation team's output. Discuss implications, limitations, and future directions.
\end{quote}

\subsection{Deliberation and synthesis prompts}

\textbf{Team deliberation} (2 rounds per phase):
\begin{quote}
\small\ttfamily
TEAM DELIBERATION --- Round \{N\}/\{max\}

Discuss your team's progress with your teammates. Consider:\\
- What you've accomplished so far\\
- Any issues or concerns about the approach\\
- Budget implications of remaining work\\
- Suggestions for your teammates

Provide a brief, constructive message (2--3 sentences).
\end{quote}

\textbf{Synthesis} (team lead produces final artifact):
\begin{quote}
\small\ttfamily
You are the \{role\_name\} (team lead) for the \{team\} team.

Your team has completed its work. Review all code outputs and discussion, then produce a single, complete artifact using write\_artifact. Do NOT write multiple versions of the same artifact.

\{Team-specific artifact template with required sections\}

CODE OUTPUTS FROM TEAM:\\
=== \{agent\_id\} ===\\
\{output (first 3000 chars)\}

TEAM DISCUSSION:\\
{[}\{agent\_id\}{]}: \{message\}
\end{quote}

\section{Simulation convergence analysis}
\label{app:convergence_sims}

\begin{figure}[t]
\centering
\includegraphics[width=\columnwidth]{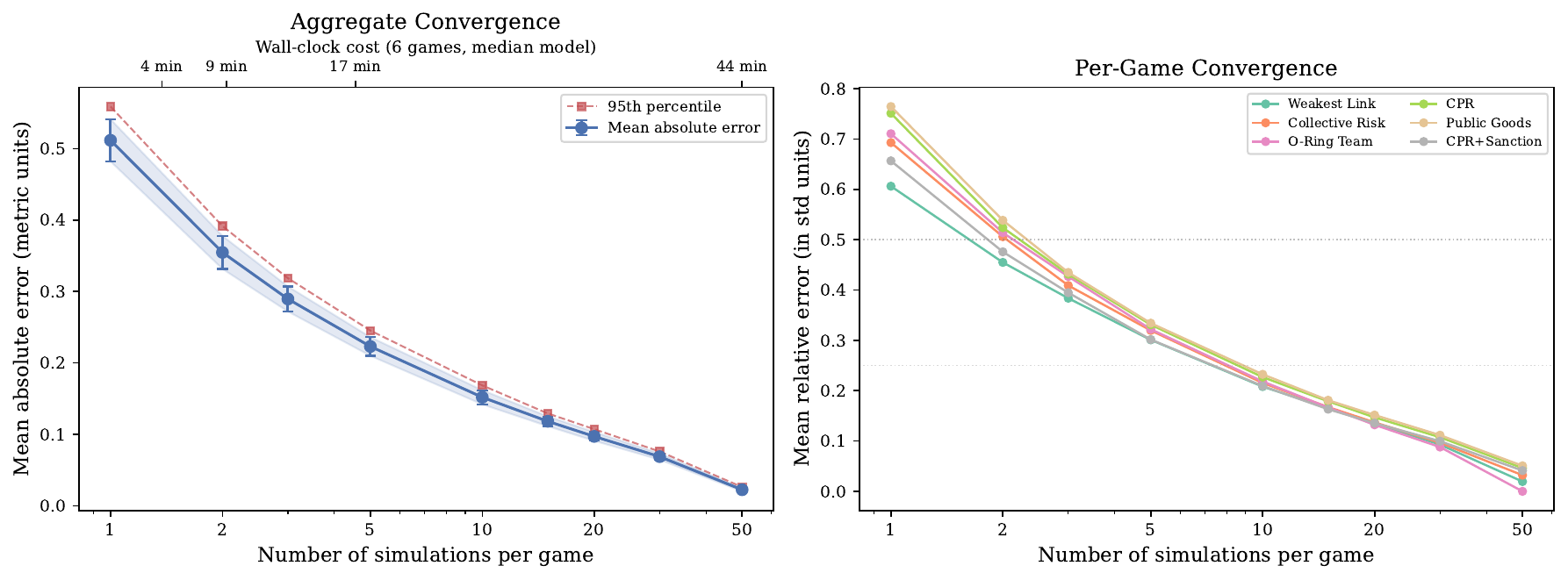}
\caption{Convergence of game metric estimates as a function of the number of simulations. \textbf{Left:} aggregate mean absolute error across all models and games (shaded region: $\pm$1 std; dashed: 95th percentile). \textbf{Right:} per-game convergence in units of the full-data standard deviation. All games show similar convergence profiles, with diminishing returns beyond 10--20 simulations.}
\label{fig:convergence}
\end{figure}

Each behavioral game configuration in Experiment 1 was run 50 times to estimate the distribution of model behaviors. However, in practice, fewer runs may be needed to estimate a model's collaborative behavior. Therefore, we compute the error for each game for our metric of choice using bootstrapped samples from 50 simulations. \fref{fig:convergence} shows the error across all six games (left) and per game (right) relative to how many total game simulations are used. Results suggest that collaborative behavior can be accurately estimated with $\sim$20 simulations, though further simulations do still lower the error. 

\section{Prompt invariance analysis}
\label{app:prompt_invariance}

To verify that cooperative behavior is not driven by surface-level prompt wording, we run a separate regression using only the canonical and alternate-prompt conditions. This is a balanced comparison where the only difference is surface-level framing (``Now make your decision'' versus\ ``Please make your decision now''). This analysis excludes CoT, ToM, and parameter sweep data to avoid confounding the prompt comparison. The prompt variant effect, shown in \tref{tab:prompt_invariance} confirms that surface-level wording has no credible effect on Pareto proximity ($\hat{\beta} = +0.007$, 95\% CI $[-0.000, +0.014]$). As a result, we use a single prompt (Standard) for all subsequent analyses.

\input{tables/tab_prompt_invariance.tex}

\section{Full Experiment 1 regression table}
\label{app:exp1_full}

Table~\ref{tab:pareto_regression_full} reports all coefficients from the Bayesian regression predicting Pareto proximity (Section~\ref{sec:games:results}), including game fixed effects and family random intercepts omitted from the main paper.

\input{tables/tab_pareto_regression_full.tex}

\section{Experiment 1 Behavioral Comparison}

\begin{figure*}[t]
\centering
\includegraphics[width=\textwidth]{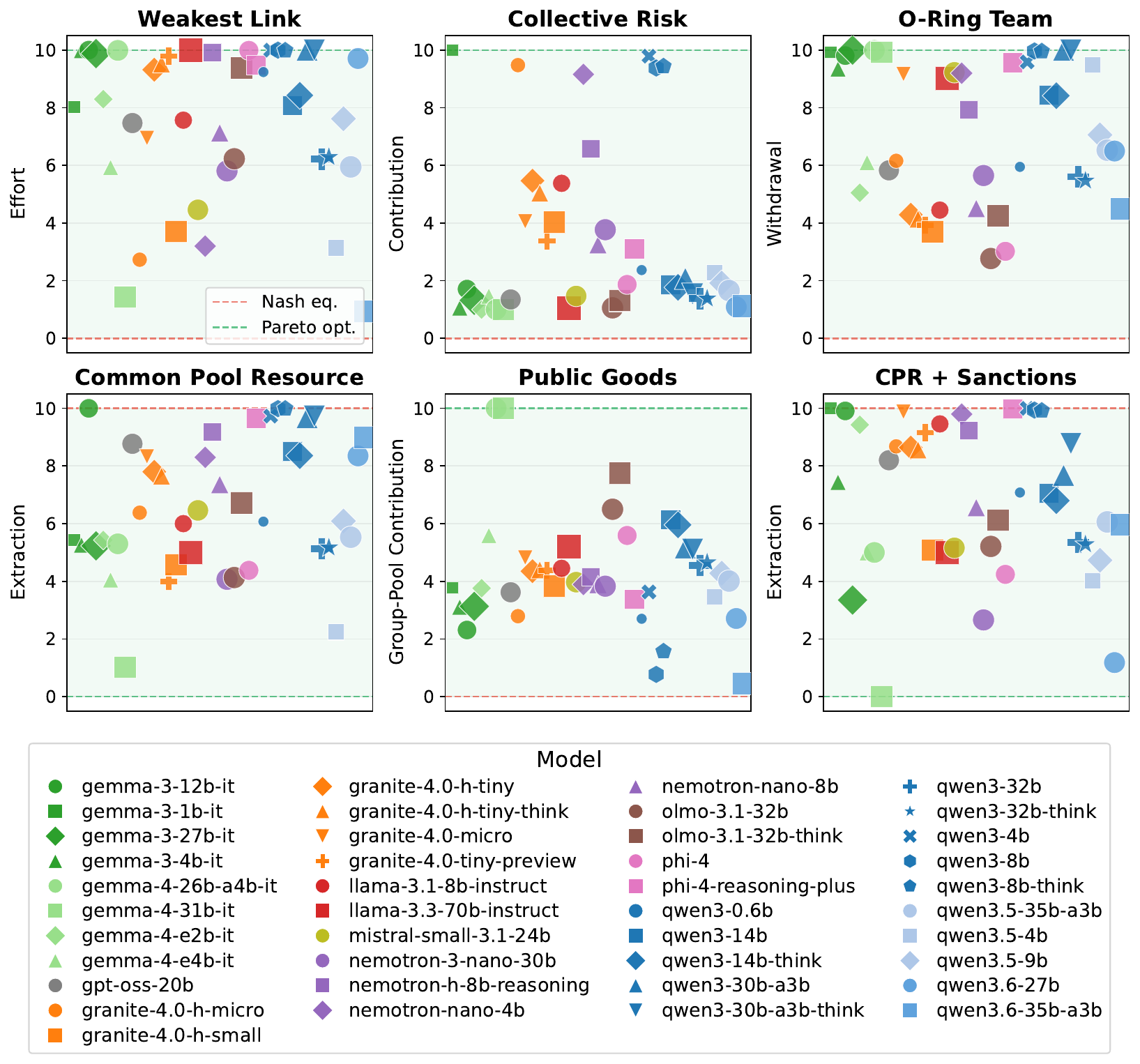}
\caption{Behavioral profiles across six games. Each subplot shows one game, with models arranged along the x-axis (sorted by family and size). The y-axis shows the primary behavioral metric. Dashed lines indicate the Nash equilibrium (red) and Pareto optimum (green); the shaded region spans the Nash--Pareto range. Point color indicates model family, marker shape distinguishes models within a family, and point size scales with $\log_2$ of the model's parameter count.}
\label{fig:nash_pareto}
\end{figure*}

Figure~\ref{fig:nash_pareto} shows behavioral profiles across all six games for 41 models. Models exhibit substantial heterogeneity. Small models (qwen3-4b, qwen3-8b, gemma-3-1b-it) tend toward high CPR extraction (9.5--10.0), near-maximum Weakest-Link effort (9.9--10.0), and excessive Collective Risk contribution (8.7--9.8 per round, well above the fair-share threshold of 1.25). These models fail to differentiate between game structures, defaulting to extreme strategies. Larger models (olmo-3.1-32b, llama-3.3-70b, phi-4) show markedly lower CPR extraction (4.1--5.3), moderate Weakest-Link effort (6.1--9.5), and near-optimal Collective Risk contribution (1.0--2.1). Most models cluster near the Pareto optimum for coordination games (Weakest-Link: 64\% within 20\% of Pareto) but scatter widely in commons dilemmas (CPR: $<$1\% near Pareto), suggesting that coordination is substantially easier for LLMs than managing shared resources.
The games show that LLMs do not exhibit a uniform pro-social bias; their cooperative tendencies are game-structure-dependent, consistent with human behavior.

\section{Full Experiment 2 regression table}
\label{app:exp2_full}

\tref{tab:regression_full} reports all coefficients from the Bayesian hierarchical measurement-error model predicting AI-for-Science outcomes (Section~\ref{sec:ailab:results}), including condition-level controls omitted from the main paper.

\input{tables/tab_ailab_regression_full.tex}

\section{Two-Stage Regression Robustness Experiment}

Our model for predicting  AI-for-Science outcomes (\tref{tab:regression}) allows the AI-for-Science likelihood to inform the fit from the latent game dispositions, which has the potential to introduce  data leakage, where the outcome variable artificially smooths the predictor. We view this risk as being low due to the number of samples per LLM, and the measurement for the latent cooperative behavior ($\theta$) is tightly specified (the per-LLM standard errors are much smaller than the cross-LLM spread of means). However, to verify this, we fit a two-stage regression, where we first estimate $\theta$ per model and then use those estimates as input to a second model for their effect on AI for Science outcomes, where we sample from $\theta$ 25 times to get an estimate $\beta$  that reflects the effect of behaviors on the outcome. All other controls and data are the same in both regression specifications.
We use a similar fitting setup as with the main regression where both stages use 4 chains $\times$ 8{,}000 samples with a 4{,}000 step warm-up.

\input{tables/tab_two_stage_games}

\tref{tab:two_stage_games} compares the parameter estimates of the joint and cut specifications of the model. The sign, magnitude, and significance are all preserved, with three exceptions: (1)  Weakest-Link's effect on Completeness which has the same sign but is no longer credible as a negative association, (2) CPR+Sanction on Accuracy, which has the same sign and magnitude but is no longer credible, and (3) CPR on Completeness, which has the same sign but becomes a credible effect. The stability of the results indicate that there is minimal forward leakage that could artificially introduce the relationships seen in our joint model.

\section{Model Inventory}
\label{app:models}

\tref{tab:models} shows the full list of 41 models used in the experiments. Models were selected to cover a diverse set of families, training methodologies, and model sizes. All models are reported (and required) to support some form of tool calling, which is necessary for the AI-for-Science experiments that use real tools to perform their workflow.

The experiments used open-weight models as a deliberate scientific choice for three reasons. First, given a fixed experimental budget for compute/API credits, we maximized the model diversity so that we could have a richer set of behavioral profiles from which we could test for connections. Second, frontier models were omitted so that our judges  (\texttt{gpt-5.4-mini} and \texttt{gemini-3.1-flash-lite-preview}) would never evaluate their own outputs. Third, open-weight models allows our experiments to be reproduced easily, have published parameter-counts we can include as controls, and let us systematically test the effect of thinking using variants on identical weights.

\input{tables/tab_models}

\section{Benchmark Scores}
\label{app:benchmark_sources}

To control for differences in model capabilities in our two experiments, we include each model's performance on three standard benchmarks. 
Table~\ref{tab:benchmark_sources} documents the source of each MMLU-Pro (5-shot, CoT), IFEval (strict prompt accuracy), and MATH-500 score used as capability controls in the regression analyses. Scores are from official tech reports or HuggingFace model cards unless noted. For thinking/instruct variant pairs sharing the same weights, MMLU-Pro scores are identical (base model capability) while IFEval may differ by mode. When publicly reported scores are not available, we compute scores using the \texttt{lm-eval} package \citep{eval-harness} using the same settings for other models. MATH-500 is unpublished for 25 of the 41 models, which we therefore scored ourselves (0-shot CoT, greedy, no tools) with a format-robust answer extractor. We additionally assessed a fourth control GPQA-Diamond, but found its scores were highly correlated with MMLU-Pro ($r$=0.87) and so it was dropped.

\input{tables/tab_benchmark_sources}

\section{AI-for-Science Token Cost}

Each AI-for-Science workflow involves multiple agents each working on specific tasks by calling tools and generating output. We quantify the cost of each experiment in terms of tokens, which are reflective of the total time required to complete the experiment and, if using APIs, the monetary cost. \tref{tab:token_usage} shows the mean number of tokens, including thinking tokens, for each model, averaged over all experiments and conditions.

\input{tables/tab_token_usage.tex}

\section{LLM-as-Judge Evaluation}
\label{app:judge_prompts}

To evaluate the AI-for-Science reports, we employ two LLM judges---one from OpenAI (\texttt{gpt-5.4-mini}) and one from Google (\texttt{gemini-3.1-flash-lite-preview})---using identical structured rubrics. Each judge independently scores every report on three metrics (completeness, quality, accuracy) on a 1--5 integer scale with one-sentence rationales. Both judges are recent and considered frontier models; neither model is used in any of our experiments, making them suitable as independent judges. Due to different rate limits, we use \texttt{gpt-5.4-mini} as the primary judge to rate all 3,473 conditions' experiment outputs and scored a subset of 1,239  overlapping conditions with \texttt{gemini-3.1-flash-lite-preview}. Judges are highly correlated in their judgments, with Pearson's $r$=0.883 for Quality, 0.830 for Completeness, and 0.590 for Accuracy. The lower correlation for accuracy is due to a difference in relative ratings; the Gemini model has a higher mean for Accuracy (0.395) than the OpenAI model (0.255), suggesting slightly different calibrations for what is considered accurate, despite using the same prompt rubric for both models. Nevertheless, we argue these high correlations support the results of Experiment 2 as being generalizable across judges. 

To ensure that GPT-5.4-mini judge faithfully mirrors human expert ratings, we collected a sample of 50 reports across science tasks, stratified by judge rating. Two experts rated the reports on Completeness, Quality and Accuracy, using the same rubric with knowledge of the task's ground truth. Annotators had high agreement on Completeness and Quality (Cohen's $\kappa$ of 0.89 and 0.79), with slightly lower agreement on Accuracy ($\kappa$=0.54); we attribute the lower agreement in Accuracy due to differences in how to assign partial credit for reports. Overall, human judges' ratings were similar to the judge used to evaluate. \texttt{gpt-5.4-mini}'s ratings had a mean absolute error from the human ground truth of 0.82, 0.68, 0.76 for Completeness, Quality and Accuracy, respectively. These results show each metric's error is well aligned, with most errors under 1 scale point away from the human ground truth.

Below we reproduce the exact system prompt and user prompt template used for all judge evaluations.

\paragraph{System prompt.}
{\small
\begin{verbatim}
You are an expert scientific reviewer evaluating reports produced by AI agent
teams. Each report was generated by a multi-agent LLM pipeline that analyzed a
dataset and produced a scientific report. You will evaluate one report at a time
on three metrics using the rubrics below.

Rate each metric on a 1-5 integer scale. Respond ONLY with valid JSON matching
the schema provided. Do not include any other text.
\end{verbatim}
}

\paragraph{User prompt template.}
{\small
\begin{verbatim}
## Task context

Science task: {task}
Task description: {task_description}

## Ground truth (known facts about the dataset)

{ground_truth_text}

## Report to evaluate

{report_text}

## Evaluation rubrics

Rate each of the following three metrics on a 1-5 scale.

### 1. Completeness (structure and coverage)

How complete is the report as a scientific document? A complete report has
seven required sections: (1) abstract, (2) introduction, (3) dataset
description, (4) methods, (5) results, (6) evaluation, (7) conclusion.
For each section, assess whether it is present with substantive content
(not just a heading or a single sentence).

- 5 (Exemplary): All 7 sections present with substantive content in each.
  Logical flow between sections.
- 4 (Good): 6 of 7 sections present with substantive content. The missing
  or thin section is minor.
- 3 (Adequate): 4-5 sections present with substantive content. The report
  reads as a partial draft.
- 2 (Poor): 2-3 sections present with substantive content. Only fragments
  of a report.
- 1 (Minimal): Fewer than 2 sections present, or the text is largely
  incoherent/boilerplate with no substantive analysis.

In addition to the overall score, report which of the 7 sections are present
with substantive content (true/false for each).

### 2. Quality (methodological rigor and presentation)

How rigorous and well-presented is the scientific analysis?

- 5 (Exemplary): Uses appropriate methodology with clear justification.
  Reports specific quantitative results (metrics, p-values, confidence
  intervals). Statistical tests are applied correctly. References upstream
  team outputs and integrates findings coherently. Tables or structured
  results present.
- 4 (Good): Sound methodology with quantitative results reported. Minor
  gaps in statistical rigor or integration of upstream work.
- 3 (Adequate): Some methodology described and some numbers reported, but
  analysis is shallow. May lack statistical tests or cross-references to
  other teams' outputs.
- 2 (Poor): Vague or boilerplate methodology. Few or no concrete numbers.
  No evidence of statistical testing or integration across pipeline stages.
- 1 (Minimal): No discernible methodology. Generic text with no
  quantitative results. Could apply to any dataset.

### 3. Accuracy (correctness of scientific findings)

How accurately does the report capture the true patterns in the data? Use
the ground truth provided above to assess this.

- 5 (Exemplary): Correctly identifies all major phenomena in the data (per
  the ground truth). Key quantitative findings are consistent with ground
  truth. Demonstrates genuine understanding of the dataset.
- 4 (Good): Identifies most major phenomena. Findings are largely
  consistent with ground truth, with minor omissions or imprecisions.
- 3 (Adequate): Identifies some true patterns but misses important ones.
  Mix of correct and incorrect or unsupported claims.
- 2 (Poor): Identifies few true patterns. Contains substantial errors or
  fabricated findings not supported by the data.
- 1 (Minimal): No correct findings. Analysis is entirely generic,
  fabricated, or contradicts the ground truth.

## Response format

Respond with ONLY this JSON object, no other text:

{
  "completeness": <int 1-5>,
  "completeness_rationale": "<one sentence>",
  "sections": {
    "abstract": <true/false>,
    "introduction": <true/false>,
    "dataset": <true/false>,
    "methods": <true/false>,
    "results": <true/false>,
    "evaluation": <true/false>,
    "conclusion": <true/false>
  },
  "quality": <int 1-5>,
  "quality_rationale": "<one sentence>",
  "accuracy": <int 1-5>,
  "accuracy_rationale": "<one sentence>"
}
\end{verbatim}
}

Each judge produces integer ratings (1--5) for completeness, quality, and accuracy, along with one-sentence rationales. For completeness, the judge additionally reports which of the seven required sections (abstract, introduction, dataset, methods, results, evaluation, conclusion) are present with substantive content. The template variables \texttt{\{task\}}, \texttt{\{task\_description\}}, \texttt{\{ground\_truth\_text\}}, and \texttt{\{report\_text\}} are filled with the specific science task name, a paragraph-length task description, formatted ground truth parameters, and the full text of the agent-generated report, respectively. Both judges use temperature 0 and a maximum of 512 output tokens.

\section{Agent Interaction Analysis}

\begin{figure}
    \centering
    \includegraphics[width=0.95\linewidth]{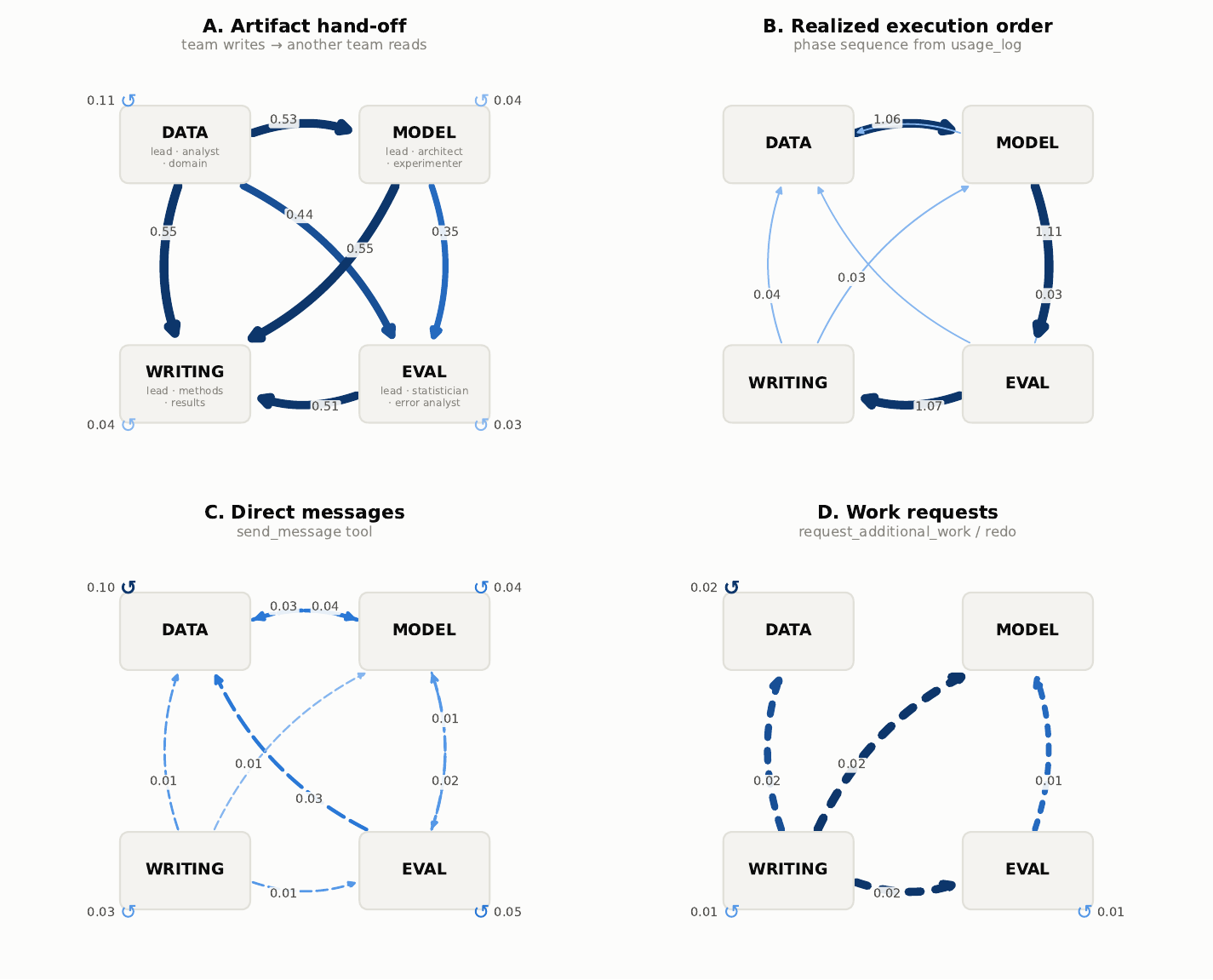}
  \caption{Inter-team interaction structure in the AI-for-Science experiments. Each node shows one team with edges for how the teams interacted. Edge weight and shade both show mean occurrences per run ($15{,}682$ runs, 41 models). The $\circlearrowleft$ loop beside a team is its intra-team self-loop. Node positions are identical across panels, so weights are comparable by position.}                         
    \label{fig:team-interaction}
\end{figure}

The agent teams in our AI-for-Science tasks interact to verify, suggest, and refine the analysis as it proceeds. These interaction steps are recorded, in part, through explicit tool calls for tasks such as sending messages to one another, reporting completed artifacts, or requesting that work be redone for a particular reason (see list in the system prompt in Appendix \ref{app:ailab_prompts}). These tools give us visibility into how teams interact. Figure \ref{fig:team-interaction} shows the mean interaction counts, averaged across all runs, faceted by specific interaction categories. We observe that teams do interact, with the most common mode being through reading and writing artifacts that other teams used or created. While relatively rare behavior, teams will also directly message to communicate or request work, suggesting agents and teams are reasoning about the larger workflow.

Although we did not find a credible benefit of ToM prompting for the quality and accuracy task outcomes (cf. \tref{tab:regression}), ToM could still be influencing the interaction structure. To test this, we examine the most common category, Artifact Handoffs, on the subset of our runs ($n$=1,159) where all conditions align except whether ToM is in the prompt or not; this subset lets us control for all other conditions influencing behavior. Within the subset, we test whether adding ToM changes the structure.

\begin{figure}
    \centering
    \includegraphics[width=0.95\linewidth]{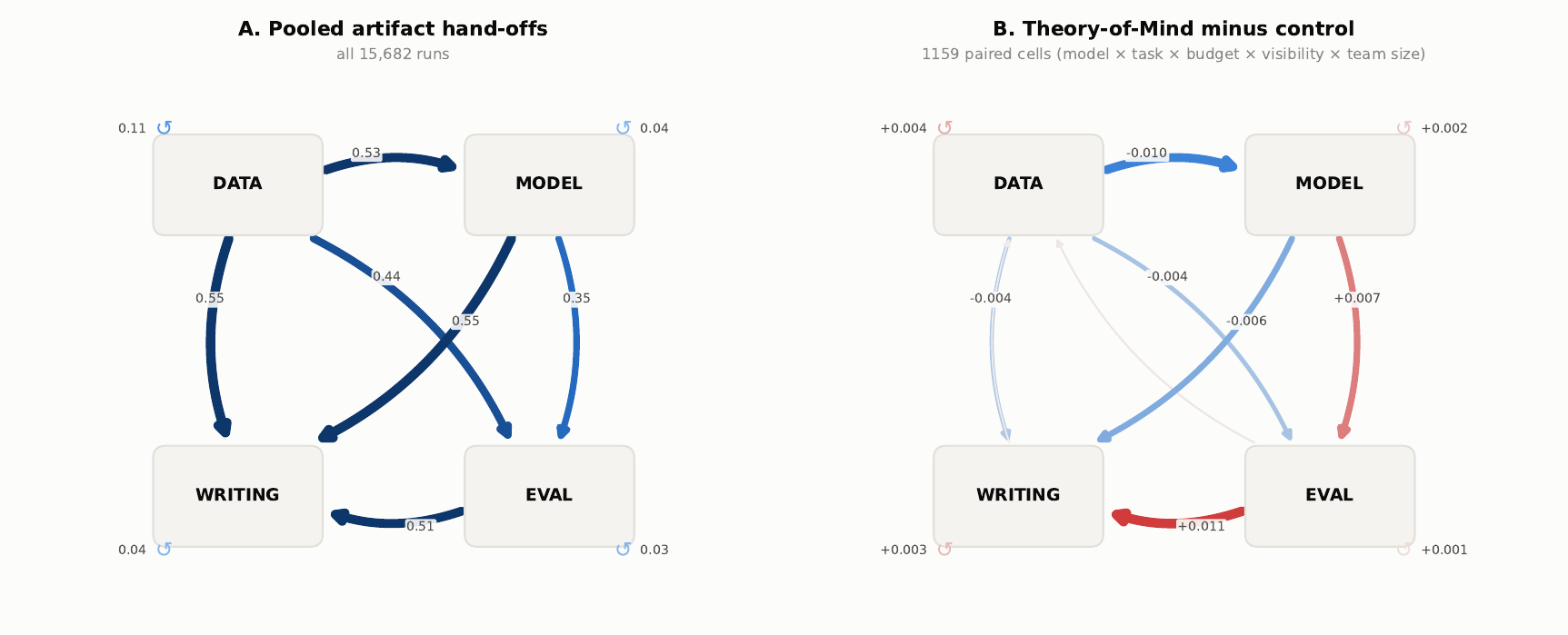}
     \caption{Theory-of-Mind (ToM) prompting does not change how work flows between teams, shown here as how often artifacts produced by one team flow to another. Panel A (left) shows the aggregate flow across all runs; Panel B (right) shows the subset of runs where all conditions aligned except whether a model was prompted to use ToM. In Panel B, red denotes that ToM increases flow on that edge, blue denotes decreases, and gray shows approximately no change. Note that because of the matching restriction, the scale is two orders of magnitude below panel A.}
    \label{fig:tom-interaction-effect}
\end{figure}

\textbf{Results}. As shown in \fref{fig:tom-interaction-effect}, ToM does not cause a substantial change in the interaction structure; no changes are significant at p$<$0.05 after multiple hypothesis testing using Holm-Bonferroni. This null result suggests that even if ToM is influencing the language of the agents and teams, ToM is not changing how these agents interact with each other. However, we emphasize that this result is specific to our experimental setting, and future LLMs used as agents may have stronger ToM abilities that do reshape the interaction patterns, so ToM should not be ruled out as an important factor.

Does a model's cooperative behavioral disposition influence the language it uses when coordinating with teammates? To test this, we curated lexicons for two categories, Cooperation (109 terms) and Self-interest (98 terms), each with subcategories for particular themes. We also remove any terms or phrases that were used in the prompt. We then fit a linear regression on rate of these terms per 1000 words, measured from the intra-team discussions and averaged across all of a model's runs. The regression is fit on Pareto distance with controls for log parameter count and whether the model is a reasoning model. Because we fit multiple models, we apply the Holm-Bonferroni correction when assessing significance.

\textbf{Results}. A model's Pareto proximity does not significantly influence the language of collaboration, as shown in \tref{tab:cooperation_language_deliberation}, though several themes approach significance at p$<$0.05. Thus, while the model's behavior in the games is predictive of their performance in realistic multi-agent settings, the language used in those settings is not substantially changed towards using more (or less) collaborative, team-based language. However, we note that our lexicon-based analysis may overlook more complex or subtle shifts in the language agents use to coordinate, and future analyses on the data we release may identify robust patterns showing how the disposition affects coordination language.

\input{tables/tab_cooperation_language_deliberation}

\end{document}

%% file: tables/tab_pareto_regression_compact.tex
\begin{table}[t]
\centering
\small
\newcommand{\grayrow}{\rowcolor{black!6}}
\sisetup{
  table-format=+1.3,
  table-number-alignment=center,
  table-space-text-post={\dag},
  input-signs=+-,
}
\begin{tabular}{l l S[table-format=+1.3] r}
\toprule
\textbf{Category} & \textbf{Predictor} & {$\hat{\beta}$} & {95\% CI} \\
\midrule
\grayrow \emph{Model}                & log(size)           & -.058 {\dag} & $[-.099,-.016]$ \\
\grayrow                             & Thinking variant    & +.021       & $[-.067,+.110]$ \\
\emph{Strategy}                      & Chain-of-thought    & +.003       & $[-.002,+.008]$ \\
                                     & Theory of mind      & -.035 {\dag} & $[-.040,-.030]$ \\
\grayrow \emph{Condition}            & Group size          & -.001       & $[-.002,+.000]$ \\
\grayrow \emph{Capability}           & MATH-500            & +.011       & $[-.050,+.073]$ \\
\emph{Game}                          & CPR                 & -.050 {\dag} & $[-.055,-.045]$ \\
\emph{\scriptsize(ref: Coll.\ Risk)}
                                     & CPR+Sanction        & -.002       & $[-.007,+.004]$ \\
                                     & O-Ring              & -.309 {\dag} & $[-.316,-.303]$ \\
                                     & Public Goods        & -.153 {\dag} & $[-.158,-.148]$ \\
                                     & Weakest-Link        & -.523 {\dag} & $[-.528,-.517]$ \\
\grayrow \emph{Variance}             & $\sigma_f$ (family) & .084       & $[.030,.163]$ \\
\grayrow                             & $\sigma$ (residual) & .222       & $[.221,.223]$ \\
\midrule
& $R^2$              & \multicolumn{2}{c}{$0.484$} \\
& $N$                & \multicolumn{2}{c}{$89{,}475$} \\
\bottomrule
\end{tabular}
\caption{Bayesian regression predicting Pareto proximity (continuous: $0{=}$Pareto, $1{=}$Nash) from model characteristics, prompting strategies, and game structure. Theory of mind prompting is the only strategy with a credible effect, reducing proximity by 0.035 (moving models closer to the social optimum). Capability is controlled with per-model MATH-500 score(s), z-scored across the cohort. Family and per-model random intercepts included (full coefficients in Appendix~\ref{app:exp1_full}). \dag\,95\% CI excludes zero. %
}
\label{tab:pareto_regression}
\end{table}

%% file: tables/tab_ailab_regression_compact.tex
\begin{table*}[t]
\centering
\newcommand{\grayrow}{\rowcolor{black!6}}
\sisetup{
  table-format=+1.3,
  table-number-alignment=center,
  table-space-text-post={\dag},
  input-signs=+-,
}
\resizebox{\textwidth}{!}{
\begin{tabular}{l l S[table-format=+1.3] r S[table-format=+1.3] r S[table-format=+1.3] r}
\toprule
& & \multicolumn{2}{c}{\textbf{Accuracy}} & \multicolumn{2}{c}{\textbf{Quality}} & \multicolumn{2}{c}{\textbf{Completion}} \\
\cmidrule(lr){3-4} \cmidrule(lr){5-6} \cmidrule(lr){7-8}
\textbf{Category} & \textbf{Predictor} & {$\hat{\beta}$} & {95\% CI} & {$\hat{\beta}$} & {95\% CI} & {$\hat{\beta}$} & {95\% CI} \\
\midrule
\grayrow \emph{Game}         & Weakest-Link     & +.019 {\dag} & $[+.008,+.030]$ & -.039 {\dag} & $[-.065,-.013]$ & -.096 {\dag} & $[-.146,-.045]$ \\
\grayrow                     & Coll.\ Risk      & +.002 {\dag} & $[+.001,+.003]$ & +.013 {\dag} & $[+.010,+.015]$ & -.023 {\dag} & $[-.027,-.019]$ \\
\grayrow                     & O-Ring           & -.023 {\dag} & $[-.028,-.017]$ & -.162 {\dag} & $[-.176,-.148]$ & -.232 {\dag} & $[-.254,-.210]$ \\
\grayrow                     & CPR              & -.065 {\dag} & $[-.082,-.047]$ & +.651 {\dag} & $[+.597,+.705]$ & +.043       & $[-.041,+.130]$ \\
\grayrow                     & CPR+Sanction     & +.020 {\dag} & $[+.001,+.039]$ & -.520 {\dag} & $[-.578,-.464]$ & +.311 {\dag} & $[+.199,+.419]$ \\
\grayrow                     & Public Goods     & -.052 {\dag} & $[-.070,-.034]$ & +.151 {\dag} & $[+.108,+.194]$ & -.268 {\dag} & $[-.342,-.195]$ \\
\emph{Benchmark}             & IFEval           & +.005 {\dag} & $[+.002,+.008]$ & +.084 {\dag} & $[+.077,+.092]$ & +.057 {\dag} & $[+.044,+.070]$ \\
                             & MMLU-Pro         & +.014 {\dag} & $[+.010,+.019]$ & -.094 {\dag} & $[-.105,-.083]$ & -.154 {\dag} & $[-.172,-.136]$ \\
                             & log(size)        & +.004       & $[-.002,+.011]$ & +.045 {\dag} & $[+.030,+.060]$ & +.081 {\dag} & $[+.055,+.108]$ \\
                             & Thinking         & -.016 {\dag} & $[-.021,-.011]$ & -.001       & $[-.013,+.011]$ & -.037 {\dag} & $[-.057,-.018]$ \\
                             & MATH-500         & +.001       & $[-.002,+.003]$ & +.020 {\dag} & $[+.014,+.027]$ & +.109 {\dag} & $[+.098,+.121]$ \\
\grayrow \emph{Condition}    & Team size = 3    & +.008 {\dag} & $[+.003,+.013]$ & +.014 {\dag} & $[+.004,+.025]$ & -.046 {\dag} & $[-.066,-.026]$ \\
\grayrow                     & Theory of mind   & +.002       & $[-.001,+.006]$ & -.001       & $[-.008,+.007]$ & +.023 {\dag} & $[+.009,+.038]$ \\
\midrule
& $\tau$ (residual SD) & \multicolumn{2}{c}{$0.06$} & \multicolumn{2}{c}{$0.11$} & \multicolumn{2}{c}{$0.20$} \\
& $N$ / Models         & \multicolumn{2}{c}{3{,}473 / 41} & \multicolumn{2}{c}{3{,}473 / 41} & \multicolumn{2}{c}{3{,}473 / 41} \\
\bottomrule
\end{tabular}
}
\caption{Bayesian hierarchical measurement-error model predicting three AI-for-Science outcomes from latent game dispositions. Game features enter as latent behavioral dispositions $\theta_{ig}$ estimated jointly with the regression across all six games. Team size is categorical (reference: 2). Additional condition controls (log budget, task dummies, visibility) included but omitted for space; full table in Appendix~\ref{app:exp2_full}. \dag\,95\% CI excludes zero. %
}
\label{tab:regression}
\end{table*}

%% file: tables/tab_prompt_invariance.tex
\begin{table}[h]
\centering
\small
\begin{tabular}{l@{\hspace{0.8em}}r@{\hspace{0.5em}}r}
\toprule
\textbf{Predictor} & $\hat{\beta}$ & 95\% CI \\
\midrule
\multicolumn{3}{l}{\emph{Model characteristics}} \\
\quad Intercept           & $+.774$\dag & $[+.648,+.901]$ \\
\quad log(size)           & $-.064$\dag & $[-.104,-.023]$ \\
\quad Thinking variant    & $+.029$      & $[-.058,+.117]$ \\
\midrule
\multicolumn{3}{l}{\emph{Capability controls (z-scored)}} \\
\quad MATH-500              & $+.011$      & $[-.049,+.071]$ \\
\midrule
\multicolumn{3}{l}{\emph{Prompt effect}} \\
\quad Standard (vs.\ alternate) & $+.007$      & $[+.000,+.014]$ \\
\midrule
\multicolumn{3}{l}{\emph{Variance components}} \\
\quad $\sigma_f$ (family)  & $.078$ & $[.022,.152]$ \\
\quad $\sigma$ (residual)  & $.232$ & $[.230,.234]$ \\
\midrule
$R^2$                     & \multicolumn{2}{c}{$0.462$} \\
$N$                       & \multicolumn{2}{c}{$25{,}925$} \\
\bottomrule
\end{tabular}
\caption{Prompt invariance test. Bayesian regression predicting Pareto proximity using only canonical and alternate-prompt conditions (no CoT, ToM, or parameter sweeps). The prompt variant coefficient is not credible ($+0.007$, 95\% CI $[-0.000, +0.014]$), confirming that standard and alternate prompt framings produce indistinguishable cooperative behavior. Game fixed effects and family random intercepts included (omitted for brevity). 4 chains $\times$ 2{,}000 samples (2{,}000 warmup); 2 divergences.}
\label{tab:prompt_invariance}
\end{table}

%% file: tables/tab_pareto_regression_full.tex
\begin{table}[h]
\centering
\small
\begin{tabular}{l@{\hspace{0.8em}}r@{\hspace{0.5em}}r}
\toprule
\textbf{Predictor} & $\hat{\beta}$ & 95\% CI \\
\midrule
\multicolumn{3}{l}{\emph{Model characteristics}} \\
\quad Intercept           & $+.833$\dag & $[+.706,+.963]$ \\
\quad log(size)           & $-.058$\dag & $[-.099,-.016]$ \\
\quad Thinking variant    & $+.021$      & $[-.067,+.110]$ \\
\midrule
\multicolumn{3}{l}{\emph{Prompting strategies}} \\
\quad Chain-of-thought    & $+.003$      & $[-.002,+.008]$ \\
\quad Theory of mind      & $-.035$\dag & $[-.040,-.030]$ \\
\midrule
\multicolumn{3}{l}{\emph{Condition controls}} \\
\quad Group size          & $-.001$      & $[-.002,+.000]$ \\
\midrule
\multicolumn{3}{l}{\emph{Capability controls (z-scored)}} \\
\quad MATH-500              & $+.011$      & $[-.050,+.073]$ \\
\midrule
\multicolumn{3}{l}{\emph{Game fixed effects (ref: Collective Risk)}} \\
\quad CPR                  & $-.050$\dag & $[-.055,-.045]$ \\
\quad CPR+Sanction         & $-.002$      & $[-.007,+.004]$ \\
\quad O-Ring               & $-.309$\dag & $[-.316,-.303]$ \\
\quad Public Goods         & $-.153$\dag & $[-.158,-.148]$ \\
\quad Weakest-Link         & $-.523$\dag & $[-.528,-.517]$ \\
\midrule
\multicolumn{3}{l}{\emph{Family random intercepts ($\hat{\delta}_f$)}} \\
\quad Gemma 3              & $-.002$      & $[-.102,+.095]$ \\
\quad Gemma 4              & $-.138$\dag & $[-.273,-.013]$ \\
\quad GPT-OSS              & $+.033$      & $[-.096,+.174]$ \\
\quad Granite 4            & $+.058$      & $[-.040,+.169]$ \\
\quad Llama                & $+.015$      & $[-.100,+.142]$ \\
\quad Mistral              & $+.031$      & $[-.099,+.182]$ \\
\quad Nemotron             & $+.052$      & $[-.043,+.155]$ \\
\quad OLMo                 & $-.059$      & $[-.187,+.052]$ \\
\quad Phi-4                & $+.018$      & $[-.095,+.134]$ \\
\quad Qwen3                & $+.041$      & $[-.048,+.133]$ \\
\quad Qwen3.5              & $-.023$      & $[-.137,+.091]$ \\
\quad Qwen3.6              & $-.014$      & $[-.135,+.102]$ \\
\midrule
\multicolumn{3}{l}{\emph{Variance components}} \\
\quad $\sigma_f$ (family)  & $.084$ & $[.030,.163]$ \\
\quad $\sigma$ (residual)  & $.222$ & $[.221,.223]$ \\
\midrule
$R^2$                     & \multicolumn{2}{c}{$0.484$} \\
$N$ / Models / Games      & \multicolumn{2}{c}{$89{,}475$ / 41 / 6} \\
\bottomrule
\end{tabular}
\caption{Full Bayesian regression predicting Pareto proximity (continuous). All coefficients shown, including game fixed effects (reference: Collective Risk) and family random intercepts. Data includes canonical, CoT, ToM, and parameter sweep conditions using the standard prompt only (see Appendix~\ref{app:prompt_invariance} for prompt invariance analysis). \dag\,95\% CI excludes zero. 4 chains $\times$ 2{,}000 samples (2{,}000 warmup); zero divergences.}
\label{tab:pareto_regression_full}
\end{table}

%% file: tables/tab_ailab_regression_full.tex
\begin{table*}[h]
\centering
\small
\begin{tabular}{l@{\hspace{1em}}r@{\hspace{0.6em}}r@{\hspace{1.5em}}r@{\hspace{0.6em}}r@{\hspace{1.5em}}r@{\hspace{0.6em}}r}
\toprule
 & \multicolumn{2}{c}{\textbf{Accuracy}} & \multicolumn{2}{c}{\textbf{Quality}} & \multicolumn{2}{c}{\textbf{Completion}} \\
\cmidrule(lr){2-3} \cmidrule(lr){4-5} \cmidrule(lr){6-7}
\textbf{Predictor} & $\hat{\beta}$ & 95\% CI & $\hat{\beta}$ & 95\% CI & $\hat{\beta}$ & 95\% CI \\
\midrule
Intercept ($\alpha$)  & $+.316$\dag & $[+.297,+.335]$ & $+.601$\dag & $[+.559,+.644]$ & $+.428$\dag & $[+.346,+.511]$ \\
\midrule
\multicolumn{7}{l}{\emph{Game behavior (latent $\theta$, z-scored)}} \\
\quad Weakest-Link       & $+.019$\dag & $[+.008,+.030]$ & $-.039$\dag & $[-.065,-.013]$ & $-.096$\dag & $[-.146,-.045]$ \\
\quad Coll.\ Risk        & $+.002$\dag & $[+.001,+.003]$ & $+.013$\dag & $[+.010,+.015]$ & $-.023$\dag & $[-.027,-.019]$ \\
\quad O-Ring             & $-.023$\dag & $[-.028,-.017]$ & $-.162$\dag & $[-.176,-.148]$ & $-.232$\dag & $[-.254,-.210]$ \\
\quad CPR                & $-.065$\dag & $[-.082,-.047]$ & $+.651$\dag & $[+.597,+.705]$ & $+.043$      & $[-.041,+.130]$ \\
\quad CPR+Sanction       & $+.020$\dag & $[+.001,+.039]$ & $-.520$\dag & $[-.578,-.464]$ & $+.311$\dag & $[+.199,+.419]$ \\
\quad Public Goods       & $-.052$\dag & $[-.070,-.034]$ & $+.151$\dag & $[+.108,+.194]$ & $-.268$\dag & $[-.342,-.195]$ \\
\midrule
\multicolumn{7}{l}{\emph{Benchmark controls (z-scored)}} \\
\quad IFEval             & $+.005$\dag & $[+.002,+.008]$ & $+.084$\dag & $[+.077,+.092]$ & $+.057$\dag & $[+.044,+.070]$ \\
\quad MMLU-Pro           & $+.014$\dag & $[+.010,+.019]$ & $-.094$\dag & $[-.105,-.083]$ & $-.154$\dag & $[-.172,-.136]$ \\
\quad log(size)          & $+.004$      & $[-.002,+.011]$ & $+.045$\dag & $[+.030,+.060]$ & $+.081$\dag & $[+.055,+.108]$ \\
\quad Thinking           & $-.016$\dag & $[-.021,-.011]$ & $-.001$      & $[-.013,+.011]$ & $-.037$\dag & $[-.057,-.018]$ \\
\quad MATH-500           & $+.001$      & $[-.002,+.003]$ & $+.020$\dag & $[+.014,+.027]$ & $+.109$\dag & $[+.098,+.121]$ \\
\midrule
\multicolumn{7}{l}{\emph{Condition controls}} \\
\quad log(budget)        & $+.001$      & $[-.001,+.004]$ & $+.003$      & $[-.002,+.008]$ & $+.001$      & $[-.009,+.010]$ \\
\quad Task: Biomarker    & $-.086$\dag & $[-.091,-.080]$ & $-.059$\dag & $[-.069,-.049]$ & $+.002$      & $[-.018,+.021]$ \\
\quad Task: Ecology      & $-.044$\dag & $[-.049,-.039]$ & $-.039$\dag & $[-.049,-.029]$ & $+.013$      & $[-.005,+.031]$ \\
\quad Task: Geospatial   & $-.058$\dag & $[-.064,-.053]$ & $-.070$\dag & $[-.080,-.060]$ & $+.014$      & $[-.005,+.033]$ \\
\quad Vis: Hidden        & $-.004$      & $[-.008,+.001]$ & $-.001$      & $[-.009,+.008]$ & $+.011$      & $[-.005,+.027]$ \\
\quad Team size = 3      & $+.008$\dag & $[+.003,+.013]$ & $+.014$\dag & $[+.004,+.025]$ & $-.046$\dag & $[-.066,-.026]$ \\
\quad Theory of mind     & $+.002$      & $[-.001,+.006]$ & $-.001$      & $[-.008,+.007]$ & $+.023$\dag & $[+.009,+.038]$ \\
\midrule
\multicolumn{7}{l}{\emph{Variance}} \\
\quad $\tau$ (residual SD)    & \multicolumn{2}{c}{$0.06$} & \multicolumn{2}{c}{$0.11$} & \multicolumn{2}{c}{$0.20$} \\
\midrule
$N$ / Models           & \multicolumn{2}{c}{3{,}473 / 41} & \multicolumn{2}{c}{3{,}473 / 41} & \multicolumn{2}{c}{3{,}473 / 41} \\
\bottomrule
\end{tabular}
\caption{Full Bayesian hierarchical measurement-error model predicting AI-for-Science outcomes, which extends Table~\ref{tab:regression} to include all condition controls. Game features enter as latent dispositions $\theta_{ig}$ across all six games. For categorical variables, the reference categories are as follows. Task reference: Anomaly. Visibility reference: Full. Team size reference: 2. \dag\,95\% CI excludes zero. 4 chains $\times$ 8{,}000 samples (4{,}000 warmup); zero divergences.}
\label{tab:regression_full}
\end{table*}

%% file: tables/tab_two_stage_games.tex
\begin{table*}[t]
\centering
\small
\begin{tabular}{l@{\hspace{2em}}r@{\hspace{0.8em}}r@{\hspace{2.5em}}r@{\hspace{0.8em}}r}
\toprule
 & \multicolumn{2}{c}{\textbf{Joint}} & \multicolumn{2}{c}{\textbf{Cut}} \\
\cmidrule(lr){2-3} \cmidrule(lr){4-5}
\textbf{Game disposition} & $\hat{\beta}$ & 95\% CI & $\hat{\beta}$ & 95\% CI \\
\midrule
\multicolumn{5}{l}{\emph{Accuracy} \quad $R^2 = .301$ (joint) $\rightarrow$ $.296$ (cut)} \\
\quad CPR              & $-.065$\dag & $[-.082,-.047]$ & $-.058$\dag & $[-.076,-.040]$ \\
\quad CPR+Sanction     & $+.020$\dag & $[+.001,+.039]$ & $+.016$ & $[-.004,+.036]$ \\
\quad O-Ring           & $-.023$\dag & $[-.028,-.017]$ & $-.022$\dag & $[-.027,-.016]$ \\
\quad Public Goods     & $-.052$\dag & $[-.070,-.034]$ & $-.046$\dag & $[-.064,-.028]$ \\
\quad Collective Risk  & $+.002$\dag & $[+.001,+.003]$ & $+.002$\dag & $[+.001,+.003]$ \\
\quad Weakest-Link     & $+.019$\dag & $[+.008,+.030]$ & $+.017$\dag & $[+.006,+.027]$ \\
\midrule
\multicolumn{5}{l}{\emph{Quality} \quad $R^2 = .533$ (joint) $\rightarrow$ $.405$ (cut)} \\
\quad CPR              & $+.651$\dag & $[+.597,+.705]$ & $+.418$\dag & $[+.378,+.460]$ \\
\quad CPR+Sanction     & $-.520$\dag & $[-.578,-.464]$ & $-.261$\dag & $[-.306,-.216]$ \\
\quad O-Ring           & $-.162$\dag & $[-.176,-.148]$ & $-.129$\dag & $[-.143,-.116]$ \\
\quad Public Goods     & $+.151$\dag & $[+.108,+.194]$ & $+.112$\dag & $[+.070,+.154]$ \\
\quad Collective Risk  & $+.013$\dag & $[+.010,+.015]$ & $+.009$\dag & $[+.007,+.012]$ \\
\quad Weakest-Link     & $-.039$\dag & $[-.065,-.013]$ & $-.051$\dag & $[-.075,-.025]$ \\
\midrule
\multicolumn{5}{l}{\emph{Completion} \quad $R^2 = .293$ (joint) $\rightarrow$ $.248$ (cut)} \\
\quad CPR              & $+.043$ & $[-.041,+.130]$ & $+.110$\dag & $[+.032,+.191]$ \\
\quad CPR+Sanction     & $+.311$\dag & $[+.199,+.419]$ & $+.119$\dag & $[+.031,+.204]$ \\
\quad O-Ring           & $-.232$\dag & $[-.254,-.210]$ & $-.214$\dag & $[-.236,-.192]$ \\
\quad Public Goods     & $-.268$\dag & $[-.342,-.195]$ & $-.172$\dag & $[-.244,-.103]$ \\
\quad Collective Risk  & $-.023$\dag & $[-.027,-.019]$ & $-.015$\dag & $[-.019,-.012]$ \\
\quad Weakest-Link     & $-.096$\dag & $[-.146,-.045]$ & $-.022$ & $[-.067,+.021]$ \\
\bottomrule
\end{tabular}
\caption{Cut-posterior robustness check for measuring AI-for-Science outcomes from behavioral dispositions. \textbf{Joint} is the model reported in the main text, in which the latent game dispositions $\theta$ and the outcome regression are estimated together. \textbf{Cut} re-fits the same specification into two stages: Stage~1 estimates $\theta$ from the behavioral game data alone, and Stage~2 regresses the outcome on draws of $\theta$ treated as fixed data, so the AI-Lab outcomes cannot influence what the behavioral measure is.}
\label{tab:two_stage_games}
\end{table*}

%% file: tables/tab_models.tex
\begin{table}[h]
\centering
\small
\setlength{\tabcolsep}{2.5pt}
\begin{tabular}{llrlrrrr}
\toprule
\textbf{Family} & \textbf{Model} & \textbf{Size} & \textbf{Type} & \textbf{MMLU-Pro} & \textbf{IFEval} & \textbf{GPQA-D} & \textbf{MATH-500} \\
\midrule
Gemma 3      & gemma-3-1b-it                  &   1B & Instruct &  14.7 &  54.5 &  19.2 &  43.4 \\
             & gemma-3-4b-it                  &   4B & Instruct &  43.6 &  90.2 &  30.8 &  70.4 \\
             & gemma-3-12b-it                 &  12B & Instruct &  60.6 &  88.9 &  40.9 &  79.2 \\
             & gemma-3-27b-it                 &  27B & Instruct &  67.5 &  90.4 &  42.4 &  83.8 \\
Gemma 4      & gemma-4-e2b-it                 &   2B & Instruct &  60.0 &  94.6 &  43.4 &  68.4 \\
             & gemma-4-e4b-it                 &   4B & Instruct &  69.4 &  96.7 &  58.6 &  73.0 \\
             & gemma-4-26b-a4b-it             &  26B & Instruct &  82.6 &  98.5 &  82.3 &  91.4 \\
             & gemma-4-31b-it                 &  31B & Instruct &  85.2 &  98.9 &  84.3 &  93.0 \\
GPT-OSS      & gpt-oss-20b                    &  21B & Instruct &  54.0 &  30.1 &  71.5 &  72.6 \\
Granite 4    & granite-4.0-h-micro            &   3B & Instruct &  43.5 &  84.3 &  25.2 &  51.2 \\
             & granite-4.0-micro              &   3B & Instruct &  44.5 &  82.3 &  27.8 &  47.4 \\
             & granite-4.0-h-tiny             &   7B & Instruct &  44.9 &  81.4 &  25.2 &  55.8 \\
             & granite-4.0-h-tiny-think       &   7B & Thinking &  44.9 &  81.4 &  25.2 &  55.8 \\
             & granite-4.0-tiny-preview       &   7B & Instruct &  29.9 &  63.0 &  31.3 &  37.6 \\
             & granite-4.0-h-small            &  32B & Instruct &  55.5 &  87.5 &  42.9 &  56.6 \\
Llama        & llama-3.1-8b-instruct          &   8B & Instruct &  48.3 &  80.4 &  30.4 &  41.8 \\
             & llama-3.3-70b-instruct         &  70B & Instruct &  68.9 &  92.1 &  50.5 &  67.8 \\
Mistral      & mistral-small-3.1-24b          &  24B & Instruct &  66.8 &  77.3 &  46.0 &  59.6 \\
Nemotron     & nemotron-nano-4b               &   4B & Instruct &  28.2 &  70.1 &  33.8 &  71.8 \\
             & nemotron-h-8b-reasoning        &   8B & Thinking &  56.4 &  71.4 &  55.1 &  94.0 \\
             & nemotron-nano-8b               &   8B & Instruct &  54.0 &  74.7 &  39.4 &  36.6 \\
             & nemotron-3-nano-30b            &  30B & Instruct &  78.3 &  83.6 &  73.0 &  82.2 \\
OLMo         & olmo-3.1-32b                   &  32B & Instruct &  49.6 &  88.8 &  48.6 &  93.4 \\
             & olmo-3.1-32b-think             &  32B & Thinking &  76.3 &  93.8 &  56.7 &  96.2 \\
Phi-4        & phi-4                          &  14B & Instruct &  70.4 &  63.0 &  56.1 &  80.4 \\
             & phi-4-reasoning-plus           &  14B & Thinking &  76.0 &  84.9 &  69.3 &  90.8 \\
Qwen3        & qwen3-0.6b                     & 0.6B & Instruct &  24.7 &  54.5 &  22.9 &  55.2 \\
             & qwen3-4b                       &   4B & Instruct &  50.6 &  81.2 &  41.7 &  84.8 \\
             & qwen3-8b                       &   8B & Instruct &  56.7 &  83.0 &  39.3 &  87.4 \\
             & qwen3-8b-think                 &   8B & Thinking &  56.7 &  85.0 &  62.0 &  97.4 \\
             & qwen3-14b                      &  14B & Instruct &  61.0 &  84.8 &  54.8 &  90.0 \\
             & qwen3-14b-think                &  14B & Thinking &  61.0 &  85.4 &  64.0 &  96.8 \\
             & qwen3-30b-a3b                  &  30B & Instruct &  78.4 &  84.7 &  54.8 &  89.8 \\
             & qwen3-30b-a3b-think            &  30B & Thinking &  78.4 &  84.7 &  65.8 &  98.0 \\
             & qwen3-32b                      &  32B & Instruct &  65.5 &  83.2 &  54.6 &  88.6 \\
             & qwen3-32b-think                &  32B & Thinking &  65.5 &  85.0 &  68.4 &  97.2 \\
Qwen3.5      & qwen3.5-4b                     &   4B & Thinking &  79.1 &  89.8 &  76.2 &  62.2 \\
             & qwen3.5-9b                     &   9B & Thinking &  82.5 &  91.5 &  81.7 &  51.2 \\
             & qwen3.5-35b-a3b                &  35B & Thinking &  85.3 &  91.9 &  84.2 &  73.0 \\
Qwen3.6      & qwen3.6-27b                    &  27B & Thinking &  86.2 &  85.4 &  87.8 &  92.8 \\
             & qwen3.6-35b-a3b                &  35B & Thinking &  85.2 &  83.9 &  86.0 &  93.0 \\
\bottomrule
\end{tabular}
\caption{Complete model inventory (41 models from 12 families). Size indicates total parameters. MMLU-Pro is 5-shot CoT; IFEval is strict prompt accuracy; GPQA-D is GPQA-Diamond (0-shot CoT); MATH-500 is the Lightman et al.~(2023) subset (0-shot CoT). All models are open-weight and served via vLLM. See Appendix for per-score source citations.}
\label{tab:models}
\end{table}

%% file: tables/tab_benchmark_sources.tex
\begin{table}[h]
\centering
\small
\setlength{\tabcolsep}{3pt}
\resizebox{\ifdim\width>\textwidth\textwidth\else\width\fi}{!}{%
\begin{tabular}{lrrrl}
\toprule
\textbf{Model} & \textbf{MMLU-Pro} & \textbf{IFEval} & \textbf{MATH-500} & \textbf{Source} \\
\midrule
\multicolumn{5}{l}{\emph{Gemma 3}} \\
gemma-3-1b-it                  &  14.7 &  54.5 & 43.4\textsuperscript{a} & \citet{gemma3techreport} T6 \\
gemma-3-4b-it                  &  43.6 &  90.2 & 70.4\textsuperscript{a} & \citet{gemma3techreport} T6 \\
gemma-3-12b-it                 &  60.6 &  88.9 & 79.2\textsuperscript{a} & \citet{gemma3techreport} T6 \\
gemma-3-27b-it                 &  67.5 &  90.4 & 83.8\textsuperscript{a} & \citet{gemma3techreport} T6 \\
\midrule
\multicolumn{5}{l}{\emph{Gemma 4}} \\
gemma-4-e2b-it                 &  60.0 &  94.6 & 68.4\textsuperscript{a} & arXiv:2607.02770 T5 \\
gemma-4-e4b-it                 &  69.4 &  96.7 & 73.0\textsuperscript{a} & arXiv:2607.02770 T5 \\
gemma-4-26b-a4b-it             &  82.6 &  98.5 & 91.4\textsuperscript{a} & arXiv:2607.02770 T5 \\
gemma-4-31b-it                 &  85.2 &  98.9 & 93.0\textsuperscript{a} & arXiv:2607.02770 T5 \\
\midrule
\multicolumn{5}{l}{\emph{GPT-OSS}} \\
gpt-oss-20b                    &  54.0 &  30.1 & 72.6\textsuperscript{c} & Our lm-eval run (not in official card) \\
\midrule
\multicolumn{5}{l}{\emph{Granite 4.0}} \\
granite-4.0-h-micro            &  43.5 &  84.3 & 51.2\textsuperscript{a} & HF model card \\
granite-4.0-micro              &  44.5 &  82.3 & 47.4\textsuperscript{a} & HF model card \\
granite-4.0-h-tiny             &  44.9 &  81.4 & 55.8\textsuperscript{a} & HF model card \\
granite-4.0-h-tiny-think       &  44.9 &  81.4 & 55.8\textsuperscript{a} & HF model card (same weights) \\
granite-4.0-tiny-preview       &  29.9 &  63.0 & 37.6\textsuperscript{a} & HF model card \\
granite-4.0-h-small            &  55.5 &  87.5 & 56.6\textsuperscript{a} & HF model card \\
\midrule
\multicolumn{5}{l}{\emph{Llama}} \\
llama-3.1-8b-instruct          &  48.3 &  80.4 & 41.8\textsuperscript{a} & \citet{llama3techreport} \\
llama-3.3-70b-instruct         &  68.9 &  92.1 & 67.8\textsuperscript{a} & HF model card \\
\midrule
\multicolumn{5}{l}{\emph{Mistral}} \\
mistral-small-3.1-24b          &  66.8 &  77.3 & 59.6\textsuperscript{c} & HF model card \\
\midrule
\multicolumn{5}{l}{\emph{Nemotron}} \\
nemotron-nano-4b               &  28.2 &  70.1 & 71.8\textsuperscript{b} & MMLU-Pro: our lm-eval run; IFEval: HF card \\
nemotron-h-8b-reasoning        &  56.4 &  71.4 & 94.0\textsuperscript{f} & arXiv:2508.14444 T3; IFEval: HF card \\
nemotron-nano-8b               &  54.0 &  74.7 & 36.6\textsuperscript{b} & MMLU-Pro: arXiv:2510.26697; IFEval: HF card \\
nemotron-3-nano-30b            &  78.3 &  83.6 & 82.2\textsuperscript{a} & HF blog; IFEval: our lm-eval run \\
\midrule
\multicolumn{5}{l}{\emph{OLMo}} \\
olmo-3.1-32b                   &  49.6 &  88.8 & 93.4\phantom{\textsuperscript{a}} & \citet{olmo31techreport} \\
olmo-3.1-32b-think             &  76.3 &  93.8 & 96.2\phantom{\textsuperscript{a}} & \citet{olmo31techreport} \\
\midrule
\multicolumn{5}{l}{\emph{Phi-4}} \\
phi-4                          &  70.4 &  63.0 & 80.4\phantom{\textsuperscript{a}} & \citet{phi4techreport} \\
phi-4-reasoning-plus           &  76.0 &  84.9 & 90.8\textsuperscript{a} & HF model card \\
\midrule
\multicolumn{5}{l}{\emph{Qwen3}} \\
qwen3-0.6b                     &  24.7 &  54.5 & 55.2\phantom{\textsuperscript{a}} & \citet{qwen3techreport} T8/T20 \\
qwen3-4b                       &  50.6 &  81.2 & 84.8\phantom{\textsuperscript{a}} & \citet{qwen3techreport} T6/T18 \\
qwen3-8b                       &  56.7 &  83.0 & 87.4\phantom{\textsuperscript{a}} & \citet{qwen3techreport} T6/T18 \\
qwen3-8b-think                 &  56.7 &  85.0 & 97.4\phantom{\textsuperscript{a}} & \citet{qwen3techreport} T6/T17 \\
qwen3-14b                      &  61.0 &  84.8 & 90.0\phantom{\textsuperscript{a}} & \citet{qwen3techreport} T5/T16 \\
qwen3-14b-think                &  61.0 &  85.4 & 96.8\phantom{\textsuperscript{a}} & \citet{qwen3techreport} T5/T15 \\
qwen3-30b-a3b                  &  78.4 &  84.7 & 89.8\textsuperscript{e} & HF model card (-Instruct-2507) \\
qwen3-30b-a3b-think            &  78.4 &  84.7 & 98.0\textsuperscript{d} & HF model card (no mode split) \\
qwen3-32b                      &  65.5 &  83.2 & 88.6\phantom{\textsuperscript{a}} & \citet{qwen3techreport} T4/T14 \\
qwen3-32b-think                &  65.5 &  85.0 & 97.2\phantom{\textsuperscript{a}} & \citet{qwen3techreport} T4/T13 \\
\midrule
\multicolumn{5}{l}{\emph{Qwen3.5}} \\
qwen3.5-4b                     &  79.1 &  89.8 & 62.2\textsuperscript{a} & HF model card \\
qwen3.5-9b                     &  82.5 &  91.5 & 51.2\textsuperscript{a} & HF model card \\
qwen3.5-35b-a3b                &  85.3 &  91.9 & 73.0\textsuperscript{a} & HF model card \\
\midrule
\multicolumn{5}{l}{\emph{Qwen3.6}} \\
qwen3.6-27b                    &  86.2 &  85.4 & 92.8\textsuperscript{a} & HF model card; IFEval: our lm-eval run \\
qwen3.6-35b-a3b                &  85.2 &  83.9 & 93.0\textsuperscript{a} & HF model card; IFEval: our lm-eval run \\
\bottomrule
\end{tabular}
}
\caption{Sources for the three benchmark scores used as capability controls in both regressions. ``T\textit{N}'' refers to Table \textit{N} of the cited paper. ``HF model card'' indicates scores taken from the model's HuggingFace repository page. The Source column gives the provenance of the MMLU-Pro and IFEval scores; MATH-500 provenance is marked on the score itself: \textsuperscript{a}\,our lm-eval run; \textsuperscript{b}\,HF card, reasoning-off; \textsuperscript{c}\,our run, custom runner; \textsuperscript{d}\,\citet{qwen3techreport} T15; \textsuperscript{e}\,\citet{qwen3techreport} T16; \textsuperscript{f}\,arXiv:2508.14444 T12, reasoning-on. A MATH-500 score with no marker is covered by the row's Source entry. }
\label{tab:benchmark_sources}
\end{table}

%% file: tables/tab_token_usage.tex
\begin{table}[t]
\centering
\small
\begin{tabular}{lr@{\hspace{0.8em}}r@{\hspace{0.8em}}r@{\hspace{0.8em}}r}
\toprule
\textbf{Model} & \textbf{Conds} & \textbf{Input/run} & \textbf{Output/run} & \textbf{Total/run} \\
\midrule
nemotron-nano-8b               &  77 &     27,745 &      5,557 &     33,302 \\
nemotron-3-nano-30b            &   8 &     21,785 &     14,058 &     35,843 \\
nemotron-h-8b-reasoning        & 127 &     58,837 &      8,909 &     67,746 \\
phi-4                          & 135 &     56,002 &     14,448 &     70,450 \\
granite-4.0-h-micro            & 136 &     57,682 &     13,984 &     71,666 \\
granite-4.0-micro              & 145 &     60,148 &     11,870 &     72,018 \\
granite-4.0-h-tiny             & 136 &     60,110 &     13,824 &     73,934 \\
gemma-3-12b-it                 & 147 &     61,287 &     13,363 &     74,650 \\
granite-4.0-h-small            & 124 &     57,721 &     17,076 &     74,797 \\
gemma-3-27b-it                 &  95 &     64,800 &     11,154 &     75,954 \\
granite-4.0-h-tiny-think       & 179 &     65,751 &     14,538 &     80,289 \\
gemma-3-4b-it                  & 141 &     70,842 &     12,906 &     83,748 \\
granite-4.0-tiny-preview       & 137 &     73,694 &     15,302 &     88,996 \\
olmo-3.1-32b                   &  92 &     61,489 &     29,654 &     91,143 \\
phi-4-reasoning-plus           &  44 &     21,685 &     91,619 &    113,304 \\
qwen3-30b-a3b                  & 129 &    104,276 &     11,406 &    115,682 \\
nemotron-nano-4b               & 106 &     67,504 &     49,692 &    117,196 \\
gemma-3-1b-it                  & 145 &    101,267 &     19,511 &    120,778 \\
olmo-3.1-32b-think             &  29 &     52,187 &     79,406 &    131,593 \\
llama-3.3-70b-instruct         & 123 &    131,017 &      8,456 &    139,473 \\
qwen3-4b                       & 107 &    110,120 &     53,885 &    164,005 \\
llama-3.1-8b-instruct          & 132 &    152,725 &     12,485 &    165,210 \\
mistral-small-3.1-24b          &  67 &    141,906 &     23,552 &    165,458 \\
qwen3-0.6b                     & 140 &    177,447 &     35,462 &    212,909 \\
qwen3-8b                       &  93 &    155,631 &     57,457 &    213,088 \\
qwen3-32b                      &  46 &    177,469 &     63,699 &    241,168 \\
qwen3-14b                      &  54 &    191,124 &     62,519 &    253,643 \\
gemma-4-e2b-it                 & 116 &    271,950 &     28,834 &    300,784 \\
qwen3-8b-think                 & 101 &    236,523 &     81,102 &    317,625 \\
qwen3-14b-think                &  59 &    260,180 &     79,625 &    339,805 \\
qwen3.5-35b-a3b                &  33 &    327,626 &     29,467 &    357,093 \\
gpt-oss-20b                    &  50 &    327,026 &     35,552 &    362,578 \\
qwen3-30b-a3b-think            &  97 &    324,050 &     39,009 &    363,059 \\
qwen3.5-9b                     &  23 &    328,774 &     45,512 &    374,286 \\
qwen3-32b-think                &  45 &    304,440 &     92,561 &    397,001 \\
qwen3.5-4b                     &  38 &    360,703 &     50,175 &    410,878 \\
gemma-4-31b-it                 &   6 &    413,651 &     40,561 &    454,212 \\
qwen3.6-35b-a3b                &   8 &    244,331 &    222,380 &    466,711 \\
qwen3.6-27b                    &  17 &    315,161 &    169,227 &    484,388 \\
gemma-4-26b-a4b-it             &  12 &    566,740 &     31,716 &    598,456 \\
gemma-4-e4b-it                 &  61 &    633,731 &     34,441 &    668,172 \\
\midrule
\emph{Median}                  &  95 &    131,017 &     29,654 &    164,005 \\
\bottomrule
\end{tabular}
\caption{AI-for-Science token usage per model, sorted by total tokens per run. Input/run and Output/run are mean tokens per simulation run; the design is 5 runs per condition, and per-run figures divide by the runs that actually completed rather than by the nominal $5\times$ conditions. Tokens include both prompt and generation tokens (including reasoning tokens for thinking models).}
\label{tab:token_usage}
\end{table}

%% file: tables/tab_cooperation_language_deliberation.tex
\begin{table}[t]
\centering
\small
\begin{tabular}{lrrcrr}
\toprule
\textbf{Category} & \textbf{Rate /1k} & $\beta$ & \textbf{95\% CI} & $R^2$ & $p$ \\
\midrule
\multicolumn{6}{l}{\emph{Cooperative Themes}} \\
Deference & 1.75 & $-5.45$ & $[-9.23, -1.67]$ & 0.29 & 0.053 \\
Credit attribution & 0.14 & $-0.43$ & $[-0.82, -0.05]$ & 0.21 & 0.187 \\
Resource stewardship & 3.57 & $+6.79$ & $[+0.63, +12.96]$ & 0.18 & 0.191 \\
Epistemic dependence & 0.11 & $-0.32$ & $[-0.69, +0.06]$ & 0.27 & 0.492 \\
Joint action & 16.38 & $+18.20$ & $[-20.73, +57.14]$ & 0.03 & 1.000 \\
Reciprocity & 0.07 & $+0.03$ & $[-0.20, +0.27]$ & 0.06 & 1.000 \\
\addlinespace
\multicolumn{6}{l}{\emph{Self-interested Themes}} \\
Credit claiming & 0.11 & $-0.50$ & $[-0.90, -0.11]$ & 0.30 & 0.106 \\
Competition & 0.06 & $+0.05$ & $[-0.15, +0.25]$ & 0.02 & 1.000 \\
Unilateral action & 0.04 & $+0.01$ & $[-0.13, +0.15]$ & 0.02 & 1.000 \\
\midrule
Cooperative index (pooled) & 22.02 & $+18.83$ & $[-24.31, +61.97]$ & 0.04 & --- \\
Self-interested index (pooled) & 0.22 & $-0.46$ & $[-1.02, +0.09]$ & 0.20 & --- \\
\bottomrule
\end{tabular}
\caption{Cooperative behavioral disposition does not predict intra-team deliberation language. Each row is a separate model-level OLS fit of the category's rate (matches per 1,000 words) on Pareto distance, controlling for $\log$ parameter count and whether the model is a reasoning model. All p-values are Holm-Bonferroni corrected. }
\label{tab:cooperation_language_deliberation}
\end{table}